\documentclass[journal]{IEEEtran}
\usepackage{algorithm}
\usepackage{algorithmic}
\usepackage{caption}

\usepackage{subcaption}
\usepackage{authblk}

\usepackage{amsmath,amssymb,graphicx,epstopdf,textpos,rotating}

\usepackage[square,sort,comma,numbers,compress]{natbib}
\usepackage{url}
\newcommand{\batch}{\mathcal{B}}

\newcommand{\expect}{\mathbb{E}}
\newtheorem{assumption}{\bf{Assumption}}
\newtheorem{lemma}{\bf{Lemma}}
\newtheorem{theorem}{\bf{Theorem}}
\allowdisplaybreaks

\ifCLASSINFOpdf
\else
\fi
\begin{document}

\title{Accelerating Minibatch Stochastic Gradient Descent using Typicality Sampling}

\author{Xinyu~Peng,~\IEEEmembership{}
        Li~Li,~\IEEEmembership{Fellow,~IEEE}%
        ~and~Fei-Yue~Wang,~\IEEEmembership{Fellow,~IEEE}
\thanks{Manuscript received ; This work was supported in part by the National Natural Science Foundation of China under Grant 61790565. \emph{(Corresponding author: Li Li.)}}
\thanks{X. Peng is with the Department of Automation, Tsinghua University, Beijing 100084, China (email: xy-peng18@mails.tsinghua.edu.cn).}
\thanks{L. Li is with the Department of Automation, Tsinghua
	National Laboratory for Information Science and Technology, Tsinghua
	University, Beijing 100084, China (email: li-li@tsinghua.edu.cn).}
\thanks{F.-Y. Wang is with the State Key Laboratory of Management and Control
	for Complex Systems, Institute of Automation, Chinese Academy of Sciences, Beijing 100080, China (e-mail: feiyue.wang@ia.ac.cn).}}

\maketitle

\begin{abstract}
Machine learning, especially deep neural networks, has been rapidly developed in fields including computer vision, speech recognition and reinforcement learning. Although Minibatch SGD is one of the most popular stochastic optimization method in training deep networks, it shows a slow convergence rate due to the large noise in gradient approximation. In this paper, we attempt to remedy this problem by building more efficient batch selection method based on typicality sampling, which reduces error of gradient estimation in conventional Minibatch SGD. We analyze convergence rate of the resulting typical batch SGD algorithm, and compare convergence properties between Minibatch SGD and the algorithm. Experimental results demonstrate that our batch selection scheme works well and more complex Minibatch SGD variants can benefit from the proposed batch selection strategy.
\end{abstract}


%
\IEEEpeerreviewmaketitle

\section{Introduction}

\IEEEPARstart{I}{n} machine learning, many proposed algorithms involve optimization of objective function requiring minimization with respect to its parameters. As an effective and popular method for large scale optimization problems, Minibatch Stochastic Gradient Descent (Minibatch SGD) exhibits the benefits from Stochastic Gradient Descent and Gradient Descent: it uses relatively few data points to approximate true gradient, which can reduce computation cost while keeping favorable convergence rate \citep{Robbins1951A}. But with rapid growth of data \citep{Krizhevsky2012ImageNet,zeiler2014visualizing,hinton2012deep,collobert2011natural} and increasing model complexity \citep{huang2017densely,ren2015faster,howard2017mobilenets} that lead to successful industrial applications, the training time by using Minibatch SGD becomes unmanageable and greatly impede research progress \citep{van2016wavenet,goyal2017accurate}.

Algorithms for accelerating Minibatch SGD have been extensively studied. Many exciting works  \citep{Qian1999On,Johnson2013Accelerating,Roux2012A,li2014efficient} focus on the update rule of model parameters, while another class of algorithms aim to increase the convergence rate by computing adaptive hyper-parameters \citep{Duchi2011Adaptive,Zeiler2012ADADELTA,Kingma2014Adam,dozat2016incorporating}. However, all of them only consider the Simple Random Sampling (SRS) scheme during the training process \citep{lohr2009sampling}, which yields an estimator of true gradient with high variance and slows down the convergence.

To this end, we propose \pmb{typicality sampling} scheme for Minibatch SGD: instead of selecting batch by SRS, we prioritize all training samples according to typicality and encourage high representative samples to be selected with greater probabilities through the whole optimization process, which helps the resulting \pmb{typical batch SGD} obtain lower gradient estimation error and guarantees faster convergence speed.

The key idea is that not all training samples are of same importance in gradient estimation. On each iteration, it only needs parts of the training set (we call them high representative samples, or typical samples) to roughly guide the correct direction of the parameters update. Increasing the chances of typical samples appearing in the selected batch can prevents the algorithm from getting trapped by low signal-to-noise ratio of gradient approximation as the iterates approach a local minimum, which improves the convergence rate, especially at the early stage of training. Our analysis shows that under certain assumptions, the proposed typical batch SGD enjoys linear convergence rate that considerably faster than Minibatch SGD. Further more, a practical implementation based on density information of training set samples and t-SNE embedding is given, which makes the proposed batch selection method more efficient. The result of improvement is empirically verified with experiments on both synthetic and natural datasets.

The rest of this paper is organized as follows. In Section \ref{section2}, we review the related literature and discuss the major drawback of existing works. In Section \ref{section3}, we outline the conventional Mini-batch SGD scheme in the context of Empirical Risk Minimization optimization, and state the basic assumptions used in this paper. In Section \ref{section4}, we introduce the details of our typicality sampling, and show the difference against SRS. In Section \ref{section5}, we theoretically prove the convergence rate of resulting typical batch SGD, and compare it with conventional Mini-batch SGD. Section \ref{section6} gives a practical implementation and discusses the experimental results on both synthetic and natural datasets. Our conclusions are presented in Section \ref{section7}.

\section{Related Work}
\label{section2}
Plenty of works have been proposed to talk about the idea that using non-uniform batch selection method for optimization process in machine learning problems. The first remarkable attempt is curriculum learning \citep{Bengio2009Curriculum}, which process the samples in an order of easiness and suggest that easy data points should be provided to the network at early stage. \cite{Kumar2010Self} propose self-paced learning that uses the loss on data to quantifies the easiness, which make the algorithm more accessible when dealing with real world dataset. However, the measurements of easiness mentioned in these two works ignore the basic spatial distribution information of training set samples, making it hard to be generalized to broader learning scenarios.

The approaches described in \cite{Loshchilov2015Online} and \cite{alain2015variance} takes advantage of importance sampling to accelerate training process. The first one proposes an online batch selection strategy that evaluates the importance by ranking all data points with respect to their latest loss value, while the latter one exhibits unbiased estimate of gradient with minimum variance by sampling proportional to the norm of the gradient. In practice, calculating the gradient norm of each sample needs a feed-forward process on all data at each iteration, which leads to quite considerable computational cost. Importance sampling with loss value may be able to alleviate this issue, but it is not a proper approximation of gradient norm.

Instead of manually designing a batch selection scheme, researchers have focused on training neural networks to select samples for the target network. The authors in \cite{jiang2017mentornet} construct a MentorNet to supervise the training process of the base deep networks through building a dynamic curriculum at each iteration. \cite{Fan2017LearningWD} propose a deep reinforcement learning framework to develop an adaptive data selection method which filters important data points automatically. Although these two approaches show promising experimental results, both of them lack solid theoretical analysis and guarantees for speedup. Moreover, the training of extra neural network is quite computationally expensive when applied to large scale dataset.

More closely related to our work, \cite{zhao2014accelerating} resort to using stratified sampling strategy for Minibatch SGD training. The authors first utilize clustering algorithm to divide training set in several groups, and then perform SRS in each group separately. This work is similar to ours, but differs significantly in two aspects: i) instead of revealing training set structure by dividing it into clusters with k-means, we apply t-SNE embedding algorithm to convert training set into low-dimensional space, which is better on capturing both local and global structure information of high-dimensional data while keeping low computational cost. ii) we do not need the corresponding label of each sample. For our approach, we distinguish typicality of each training sample by its contribution to the true gradient, which is then transformed into the form of density information in practical implementation.

Compared to aforementioned works, our proposed typicality sampling scheme has several potential advantages. First, by letting high representative samples dominant the computation of search direction, our algorithm enables more accurate estimation of the true gradient, which leads to faster convergence speed. Second, our implementation is straightforward with negligible additional computational requirement, and based entirely on the density information of training set samples without the use of expert knowledge. Third, our method has strong extensibility and can be coupled with other variant Minibatch SGD algorithms, improving the convergence performance while keeping the advantages of original algorithms unchanged.

\section{Problem Presentation}
\label{section3}
Let $f (x_i ;\theta)$ be the output predicted by the model with parameters $\theta$ when data observation drawn from training set $\mathcal{X}$ is $x_i$. Our goal is to find a approximate solution of the Empirical Risk Minimization (ERM) problem, which is often typified as the form
\begin{equation}
\min_{\theta \in \Theta} J(\theta) =  \frac{1}{N}\sum_{i=1}^N \ell(f (x_i ;\theta),y_i).
\label{ori-obj}
\end{equation}
In general case, each term $\ell(\cdot)$ models the loss between predicted result and corresponding label $y_i$. $J(\theta)$ measures the overall misfit across the whole training set. As a popular method for minimizing \eqref{ori-obj}, Minibatch SGD usually performs the following update rule on iteration $k$
\begin{eqnarray}
\label{mini-batch sgd update rule}
\theta_{k+1} &=& \theta_k - \eta_k\nabla J_\batch (\theta),\nonumber\\
\text{ where } \,\,\, \nabla J_\batch (\theta) &=& \frac{1}{m}\sum_{i \in\batch}\nabla\ell(f (x_i ;\theta),y_i),
\end{eqnarray}
and $\batch$ is a mini-batch sampled from $\mathcal{X}$ using SRS method, where $m$ is the Minibatch size and $\eta_k$ is the learning rate. Based on \eqref{mini-batch sgd update rule}, we can regard the stochastic gradient $\nabla J_\batch (\theta)$ constructed in Minibatch SGD as a noisy estimation of true gradient $\nabla J(\theta_k)$. Thus we define the search direction on iteration $k$ as
\begin{equation}
\label{noisy gradient}
\nabla J_\batch (\theta_k) := \nabla J(\theta_k) + e_k,
\end{equation}
where $e_k$ denotes the residual term. Although $\nabla J_\batch (\theta)$ is an unbiased estimator of $\nabla J(\theta_k)$, the inaccurate estimation, especially the large value of $\expect \Vert e_k\Vert^2$ bringing by SRS, still affects the convergence rate of Minibatch SGD (See Section \ref{section4} for details).

For simplicity, we use the following notations throughout the remainder of this paper. By $\theta_k$, we denote the value of parameter $\theta$ on iteration $k$. By $\theta_*$, we denote a global minimizer of $J(\theta)$. We use $\nabla J_i(\theta)$ to denote $\nabla\ell(f (x_i ;\theta),y_i)$. We also make the following three common assumptions.
\begin{assumption}[\cite{Hiriarturruty1993Convex}]
	\label{assumption1_lipschitz}
	We assume that a minimizer $\theta_*$ always exists, that $J_i(\theta)$ are differentiable, and that $\nabla J(\theta)$ is Lipschitz continuous, i.e., for any $x,y \in \Theta$ and some positive L:
	\begin{equation}
	\label{inequlity_lipschitz}
	\| \nabla J(x) - \nabla J(y) \| \le L \| x - y \|.
	\nonumber
	\end{equation}
\end{assumption}
This is a standard smoothness assumption in the optimization literature \citep{Heinonen2005LECTURES,Friedlander2011Erratum,Csiba2016Importance}. Note that as a consequence of Assumption \ref{assumption1_lipschitz}, we have following inequality holds:
\begin{equation}
\label{lipschitz_L_inequality}
J(y) \le J\left( x \right)+{{\left( y-x \right)}^{T}}\nabla J\left( x \right)+\frac{L}{2}{\| y-x \|}^{2}
\end{equation}

\begin{assumption}[\cite{Nesterov2004Introductory}]
	\label{assumption2_stronglyconvex}
	We assume that the objective function $J(\theta)$ is strongly convex with positive parameter $\mu$, i.e., for any $x,y \in \Theta$, $J(\theta)$ satisfies:
	\begin{equation}
	\label{stronglyconvex_lambda_inequality}
	J(y) \ge J(x)+{(y-x)^{T}}\nabla J(x)+\frac{\mu}{2}{{\left| \left| y-x \right| \right|}^{2}}.
	\end{equation}
\end{assumption}
This is another assumption usually used in optimization theory \citep{Boyd2006Convex,Duchi2011Adaptive,De2016Big}. Under this assumption, the ERM problem \eqref{ori-obj} becomes a strongly convex programming problem.
\begin{assumption}[\citep{Bertsekas1996Neuro}]
	\label{assumption3_stochistic_ek}
	We assume that for some constants $\beta_1\ge0$ and $\beta_2\ge1$, the following inequality holds for any $\theta \in \Theta$:
	\begin{equation}
	{{\|\nabla {{J}_{i}}(\theta)\|}^{2}}\le {{\beta}_{1}}+{{\beta}_{2}}{{\|\nabla J(\theta)\|}^{2}} \,\,\,\,\,\,\,\,\, i=1,...N.
	\nonumber
	\end{equation}
\end{assumption}
This is a standard assumption in stochastic optimization theory \citep{Friedlander2011Erratum}, which implies functions $\nabla J_i(\theta)$ are bounded by the true gradient $\nabla J(\theta)$. Besides above three assumptions, our analysis is also based on the following lemma, which explain the relationship between batch selection scheme and convergence speed.
\begin{lemma}[\cite{Friedlander2011Erratum}]
	\label{lemma1}
	Given learning rate $\eta \equiv 1/L$, suppose Assumption \ref{assumption1_lipschitz} and Assumption \ref{assumption2_stronglyconvex} hold. At each iteration $k$, we have the following bound for those algorithms using updates of the form \eqref{mini-batch sgd update rule}:
	\begin{equation}
	E[J({{\theta}_{k+1}})-J({{\theta }_{*}})] \le (1-\frac{\mu}{L})E[(J({{\theta }_{k}})-J({{\theta }_{*}}))] +\frac{1}{2L}E[{{\|{{e}_{k}}\|}^{2}}].
	\end{equation}
\end{lemma}
Clearly, different algorithm in which an update rule of form \eqref{mini-batch sgd update rule} is applied will convergences at a different rate that determined by $\expect \Vert e_k\Vert^2$. Smaller the expected term is, faster the algorithm convergences. Therefore, batch selection method, as an important part of the algorithms to determine the expected value of gradient error, will directly affect the convergence speed.

\section{Two Batch Selection Method}
\label{section4}

\subsection{Simple Random Sampling}
SRS is one of the most basic sampling scheme that often used in solving finite sampling problem. It provides theoretical support for more complicate form of probability sampling method. A sample of size $m$ drawn by SRS has $m$ distinct units so that every subset of $m$ unique samples has the same probability to be selected. It follows from this definition that the probability of selecting any sample S of size $m$ is $P(S)={1}/{\binom{N}{m}}$. So each unit is of same probability $\pi$ to be selected in sample $S$
\begin{equation}
\label{standard_sampling_rate}
\pi=\frac{m}{N}
\end{equation}
Conventional Minibatch SGD usually chooses SRS as the method for generating batch when computing search direction. In this case, a batch $\batch \subset \mathcal{X}$ of data is sampled uniformly at random on each iteration, and the gradient estimate is then constructed based on the samples in selected batch $\batch$. The standard Minibatch SGD can be summarized as Algorithm \ref{Mini-batch_SGD}. As a consequence of this algorithm, authors in \cite{lohr2009sampling} prove the following result of $E{{\|{{e}_{k}}\|}^{2}}$.
\begin{lemma}
	\label{lemma2}
	\citep{lohr2009sampling}
	Suppose we use SRS as batch selection method of Minibatch SGD, then we have following result at each iteration $k$:
	\begin{equation}
	\label{mini-batch_ek}
	E{{\|{{e}_{k}}\|}^{2}} = (1-\frac{m}{N})\frac{\mathcal{S}_{k}^{2}}{m},
	\end{equation}
	Where $\mathcal{S}_{k}^{2}=\frac{\sum_{i=1}^N {{\|\nabla{{J}_{i}}({{\theta }_{k}})-\nabla J({{\theta}_{k}})\|}^{2}}}{ N-1}$.
\end{lemma}
In conventional Minibatch SGD, all samples in the training set are considered to be of same typicality when estimating stochastic gradient, which leads to the result shown in Lemma \ref{lemma2}: each data point contributes to the calculation of gradient error in expectation with same weight, and the optimization progresses inefficiently with large gradient error.

\begin{algorithm}[]
	\begin{algorithmic}[1]
		\REQUIRE Global learning rate $\eta$
		\REQUIRE Batch size $m$
		\REQUIRE Training set $\mathcal{X}=\{x_1,x_2,......x_n\}$
		\REQUIRE Initial model parameter $\theta_{0}$
		\WHILE {$\theta_{k}$ not converged}
		\STATE Update iteration: $k \gets k+1$
		\STATE Get batch: select batch $\batch$ of size $m$ from training set $\mathcal{X}$ by SRS
		\STATE Compute gradient : $\nabla J_\batch (\theta_k)=\sum_{i\in \batch}\nabla {{J}_{i}}(\theta)/m$
		\STATE Apply update: $\theta_{k+1}=\theta_{k}-\eta*\nabla J_\batch (\theta_k)$    
		\ENDWHILE
	\end{algorithmic} 
	\caption{Minibatch SGD}  
	\label{Mini-batch_SGD}
\end{algorithm}

\subsection{Typicality Sampling}
\label{subsection_42}
Drawing inspiration from the drawback of SRS, we design a more efficient sampling scheme by using additional information, e.g., the typicality of training set data. The idea behind is that: we should pay more attention to the samples which can reflect the key pattern of the hypothesis space employed for the given application, especially at the early stage of training. Thus, training samples should be considered of different importance. As the most important part of the training set in true gradient estimation, the typical samples can generate more accurate search direction by providing most informative gradient features. To this end, we first make the following assumption.
\begin{assumption}
	\label{assumption4_trainingset}
	We assume that the overall gradients of high representative subset $\mathcal{H}$, which consists of all typical samples in training set $\mathcal{X}$, satisfy:
	\begin{align}
	\sum_{i\in \mathcal{H}}\nabla {{J}_{i}}\left( \theta  \right) = \nabla J\left( \theta  \right)\,\,\,\,\,\, \text{for any} \,\,\,\,\,\theta \in \Theta.
	\nonumber
	\end{align}
\end{assumption}
By this assumption, we show that the gradients on typical samples in high representative subset $\mathcal{H}$ can reflect the correct update direction of model parameters, which suggests that we should assign higher weights to the typical samples to speed up the convergence. 

In our proposed \pmb{typicality sampling} scheme, we achieve this idea by encouraging samples in subset $\mathcal{H}$ to be selected with greater probabilities, which then make the high representative samples achieve a dominant position in computing search direction, and reduce error. A small part of batch $\batch$ are selected from the rest samples to consider the training set comprehensively. Specifically, the typicality sampling scheme performs the following steps:

\begin{enumerate}
	\item  According to the conditions in Assumption \ref{assumption4_trainingset}, demarcate subset $\mathcal{H}$ with size $N_1$ from training set $\mathcal{X}$ and group the rest samples into subset $\mathcal{L}$ with size $N_2$.
	\item  To make typical samples be selected with greater probabilities, we choose a (positive integer) parameter $n_1$ large enough that
	\begin{equation}\label{size_of_n_1}
	\frac{n_1}{N_1} \ge \frac{n_2}{N_2},
	\end{equation} 
	Where $n_1 \in (0,m)$ and $n_2=m-n_1$. Then perform step 3 at each iteration $k$.
	\item  Combine sub-batch $H_k$ and sub-batch $\ell_k$ to get batch $\batch$, where $H_k$ with size $n_1$ and $\ell_k$ with size $n_2$ are sampled from subset $\mathcal{H}$ and subset $\mathcal{L}$ by SRS respectively.
\end{enumerate}

Based on the above definition, the Minibatch SGD with typicality sampling scheme, which we call \pmb{typical batch} \pmb{SGD}, can be summarized as Algorithm \ref{typical batch SGD}. Similar with Lemma \ref{lemma2}, we have following result of $E{{\|{{e}_{k}}\|}^{2}}$ in the case of typical batch SGD.
\begin{lemma}
	\label{lemma3}
	Let us define $\beta=(n_1  N)/(m  N_1)$. Suppose training set $\mathcal{X}$ satisfies Assumption \ref{assumption4_trainingset}, then we have following result of the typical batch SGD:
	\begin{eqnarray}
	\label{new_batch_ek}
	E{{\|{{e}_{k}} \|}^{2}}&=&{{\Big\|\big(\beta-1\big)\cdot\nabla J( {{\theta }_{k}})\Big\|}^{2}}+(1-\frac{{n}_{1}}{{{N}_{1}}})\frac{n_1}{m^2}\mathcal{S}_{H,k}^{2} \nonumber\\
	& &+(1-\frac{n_2}{{{N}_{2}}} )\frac{n_2}{m^2}\mathcal{S}_{\ell,k}^{2},
	\end{eqnarray}
	Where $\mathcal{S}_{H,k}^{2}=\frac{\sum_{i\in \mathcal{H}}{{\|\nabla {{J}_{i}}( {{\theta }_{k}})-\nabla J( {{\theta }_{k}})\|}^{2}}}{{{N}_{1}}-1}$ and $\mathcal{S}_{\ell,k}^{2}=\frac{\sum_{i\in \mathcal{L}}{{\|\nabla {{J}_{i}}( {{\theta }_{k}})\|}^{2}}}{{{N}_{2}}-1 }$.
\end{lemma}  
The proof is given in the appendix. In typical batch SGD, the independence of subset $\mathcal{H}$ and subset $\mathcal{L}$ allow us to treat data points unequally and perform non-uniform sampling scheme respectively. Samples from high representative subset will contribute to the calculation of $E{{\|{{e}_{k}}\|}^{2}}$ with higher weight if \eqref{size_of_n_1} holds on each iteration k, which leads to the result in above lemma. In the succeeding sections, we demonstrate the convergence speedup of our proposed algorithm by both providing theoretical guarantees and experimenting with synthetic and natural datasets. 

\begin{algorithm}[]
	\begin{algorithmic}[1]
		\REQUIRE Global learning rate $\eta$
		\REQUIRE Batch size $m$
		\REQUIRE Training set $\mathcal{X}=\{x_1,x_2,......x_n\}$
		\REQUIRE Initial model parameter $\theta_{0}$
		\REQUIRE Batch selection rate $\gamma \in (0,0.5)$
		\STATE Build subset: demarcate subset $\mathcal{H}$ from training set $\mathcal{X}$ and put the other samples in subset $\mathcal{L}$ 
		\WHILE {$\theta_{k}$ not converged}
		\STATE Update iteration: $k \gets k+1$
		\STATE Select sub-batch $\mathcal{H}_{k}$ of size $n_1$ from subset $\mathcal{H}$ by SRS
		\STATE Select sub-batch $\mathcal{L}_{k}$ of size $n_2$ from subset $\mathcal{L}$ by SRS
		\STATE Get batch: $\batch \gets \mathcal{H}_{k}+\mathcal{L}_{k}$
		\STATE Compute gradient : $\nabla J_\batch (\theta_k)=\sum_{i\in \batch}\nabla {{J}_{i}}(\theta)/m$
		\STATE Apply update: $\theta_{k+1}=\theta_{k}-\eta*\nabla J_\batch (\theta_k)$    
		\ENDWHILE
	\end{algorithmic} 
	\caption{Typical batch SGD}  
	\label{typical batch SGD}
\end{algorithm}

\section{Convergence Analysis}
\label{section5}
\begin{figure}[t]
	\begin{center}
	\includegraphics[scale=0.4]{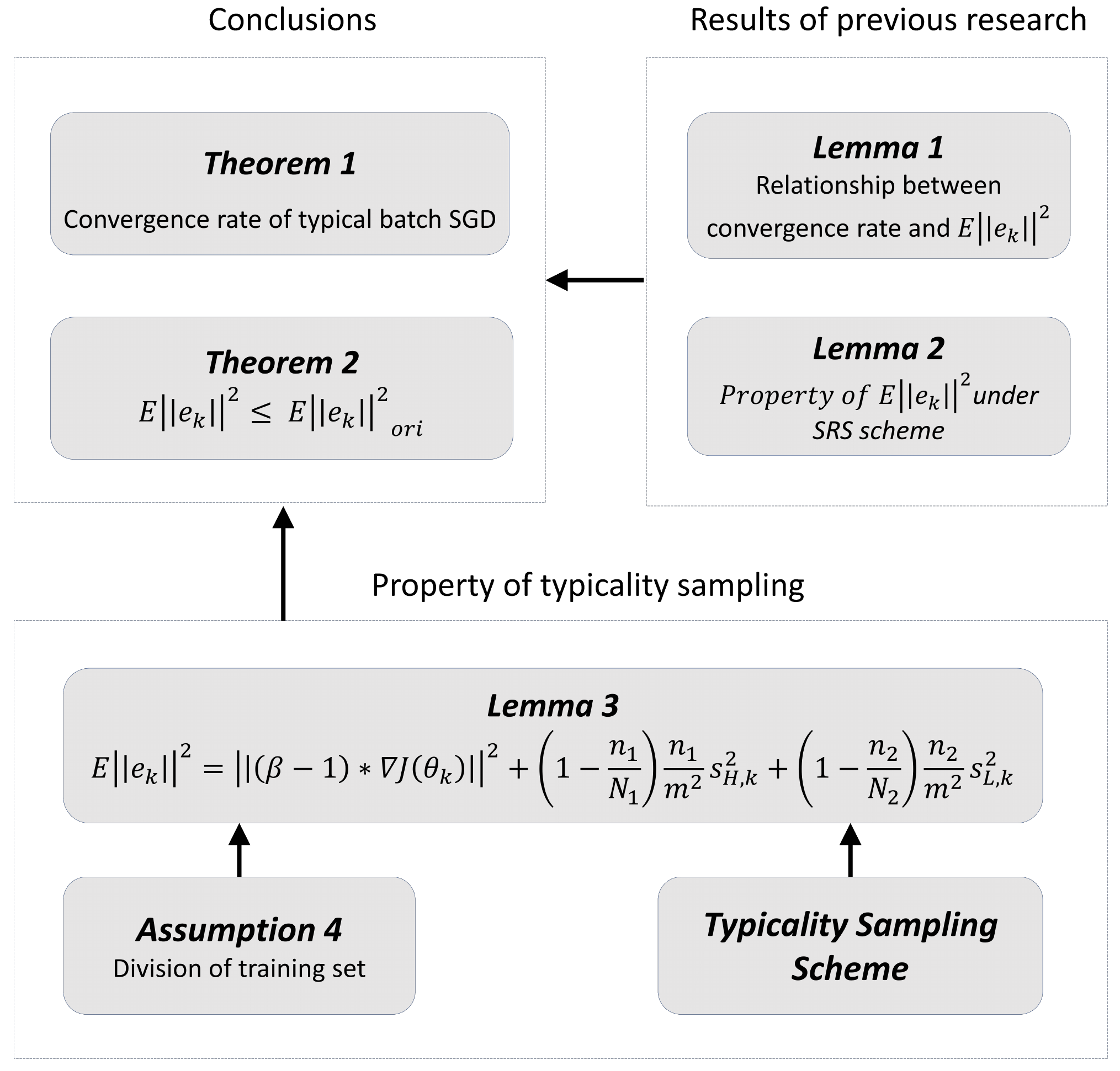}
	\caption{Structure of the proof. The property of proposed typicality sampling scheme, together with the results of previous research leads to the conclusions.}
	\label{structure_of_proof}
    \end{center}
\end{figure}

In this section, we give a convergence analysis of proposed algorithm and compare it against conventional Minibatch SGD. Our main conclusion is drawn in Theorem \ref{convergence_rate} and Theorem \ref{compare_with_sgd}. The logic flow of the whole proof is illustrated in Figure \ref{structure_of_proof}.

Using Lemma \ref{lemma1} and Lemma \ref{lemma3}, we now first provide convergence rate for typical batch SGD.
\begin{theorem}
	\label{convergence_rate}
	Suppose that Assumption \ref{assumption1_lipschitz}, \ref{assumption2_stronglyconvex} and \ref{assumption3_stochistic_ek} hold, and training set satisfies Assumption \ref{assumption4_trainingset}. Suppose further that $n_1$ is large enough so that \eqref{size_of_n_1} holds on each iteration k, then we have following convergence rate for the proposed typical batch SGD that uses update rule \eqref{mini-batch sgd update rule}:
	\begin{align}
	\label{new_convergence_rate}
	\begin{split}
	E&[J({{\theta }_{k+1}} )-J( {{\theta }_{*}} ) ]\le \Big[ 1-\frac{\mu }{L}+( 1-\frac{n_1}{{{N}_{1}}} )\frac{2n_1(\beta_{2}+2)}{m^2}\\&+( 1-\frac{n_2}{{{N}_{2}}} )\frac{n_2(\beta_{2}+1)}{2m^2}+{{( \beta -1 )}^{2}} \Big]*E[ ( J( {{\theta }_{k}} )-J( {{\theta }_{*}} ) ) ],
	\end{split}
	\end{align}
	Where $\eta \equiv 1/L$ and $\beta=(n_1  N)/(m  N_1)$. Assume that $m$ is sufficiently large so that 
	\begin{align}
	\left[( 1-\frac{n_1}{{{N}_{1}}} )\frac{2n_1(\beta_{2}+2)}{m^2}+( 1-\frac{n_2}{{{N}_{2}}} )\frac{n_2(\beta_{2}+1)}{2m^2}+{{( \beta -1 )}^{2}}\right] < \frac{\mu }{L} \nonumber
	\end{align}
	then we have linear convergence rate in expectation for our typical batch SGD. 
\end{theorem}
The proof is in the appendix. Note that our Theorem \ref{new_convergence_rate} implies the batch size $m$ affects the final convergence rate greatly. This observation matches previous analysis, since typical batch SGD can expect more benefits from the proposed sampling scheme with more typical samples introduced to batch $\batch$.
\subsection{Comparison to Minibatch SGD}
We now show that the proposed typical batch SGD exhibits faster convergence rate against conventional Minibatch SGD. The theoretical analysis is based on the observation of Lemma \ref{lemma1} that convergence rate is determined by the expected value of gradient error, with smaller expected term corresponding to faster convergence. Therefore, to see effect of the new batch selection method, we should compare ${\expect\|e_k\|^2}$ in typical batch SGD with ${\expect\|e_k\|^2}_{ori}$ in Minibatch SGD. The following result reveals the relationship between these two terms.\\
\begin{theorem}
	\label{compare_with_sgd}
	Suppose that Assumption \ref{assumption3_stochistic_ek} holds and training set satisfies Assumption \ref{assumption4_trainingset}. Suppose further that $n_1$ is large enough so that \eqref{size_of_n_1} holds on each iteration k, then we get:
	\begin{equation}\label{e_k_relationship}
	E{{\| {{e}_{k}} \|}^{2}}\le E{{\| {{e}_{k}} \|}^{2}}_{ori}.
	\end{equation}
\end{theorem}
Theorem \ref{compare_with_sgd} shows that the typicality sampling scheme is more efficient than SRS by reducing error of gradient estimation, which then improves the convergence of Minibatch SGD. The benefits are intuitive: the better use of the typicality information of training set data compensate the loss of unbiased estimator, and help the search direction becomes more accurate as typical samples dominate batch $\batch$. To quantify the degree of improvement, we define parameter $\alpha$ according to:$$E{{\| {{e}_{k}} \|}^{2}}:=\alpha E{{\| {{e}_{k}} \|}^{2}}_{ori}$$Clearly that smaller $\alpha$ implies greater improvement. Note that in practice, to achieve minimal value of $\alpha$ which represents that the proposed sampling scheme is most effective, we take $\beta={1}/(2{\sqrt[3]{\frac{m}{4}+\sqrt{\frac{{{m}^{3}}}{27}}}+\sqrt[3]{\frac{m}{4}- \sqrt{\frac{{{m}^{3}}}{27}}}})$ and set $n_1=0.8m$. 

\section{Practical Implementation and Experiments}
\label{section6}

\subsection{Practical Implementation with t-SNE}
From algorithm \ref{typical batch SGD}, we see that the preprocessing step, in which high representative subset $\mathcal{H}$ is demarcated from training set accordingly, is of critical importance to the proposed algorithm. While one can strictly follow the conditions in Assumption \ref{assumption4_trainingset} to partition the training set at each iteration, the purpose of typical batch SGD is to enable us to train deep networks with high efficiency, while the above method can be very time-consuming. In this section, we show that this issue can be alleviated by a more feasible method that using the density information of training set, and the resulting subsets approximately satisfy the assumption. 

It follows from the analysis of Assumption \ref{assumption4_trainingset} that the gradient on typical samples in subset $\mathcal{H}$ is almost the same of true gradient. Basing on this observation, we compare high density data points in density-based clustering algorithm and high representative samples in typical batch SGD. Refer to the definition in \cite{tan2013data}, data points can be classified as being in high dense regions (include core points and border points) or in sparse regions (include noise points). It only needs the high density samples to represent the whole dataset: core points found clusters and border points fill them. To mimic the effect on gradient estimation, we assume that the samples which represent the whole training set can also roughly generate the true gradient. Thus when the overall noise is gaussian white noise and large datasets are used, we have
\begin{gather}
\underset{i\in High\_d}{\mathop \sum }\nabla {{J}_{i}}\left( \theta  \right) = \nabla J\left( \theta  \right) - \underset{i\in Low\_d}{\mathop \sum }\nabla {{J}_{i}}\left( \theta  \right) = \nabla J\left( \theta  \right) \nonumber
\end{gather}
where $High\_d$ denotes the subset of all samples from highly dense regions. The above results shows that in the proposed algorithm, high density subset $High\_d$ is functional equivalent to high representative subset $\mathcal{H}$. Dividing the training set according to density information satisfies the conditions in Assumption \ref{assumption4_trainingset}.

In practice, we first apply t-SNE algorithm to embed high-dimensional training set to 2 dimensions, and utilize gaussian kernel density estimation algorithm to compute density values of all training samples in the low dimensional space, which are then used to partition the training set. On each iteration $k$, we fix subset size $n_1=0.8m, n_2=0.2m$ to achieve the best improvement. We also set $N_1 \le N_2$ to keep \eqref{size_of_n_1} holds. The complete details are listed in Algorithm \ref{typical batch SGD tsne}.

\begin{algorithm}[t]
	\begin{algorithmic}[1]
		\REQUIRE Global learning rate $\eta$
		\REQUIRE Batch size $m$
		\REQUIRE Training set $\mathcal{X}=\{x_1,x_2,......x_n\}$
		\REQUIRE Initial model parameter $\theta_{0}$
		\REQUIRE Batch selection rate $\gamma \in (0,0.8)$
		\STATE Compute 2-dimensional data representation by t-SNE: $\mathcal{X}'=\{x_1',x_2',......x_n'\}$
		\STATE Compute the density of each sample in $\mathcal{X}'$: $\mathcal{D}=\{d_{x_1'},d_{x_2'},......d_{x_n'}\}$
		\STATE Build subset $\mathcal{H}$: select the top $n*\gamma$ samples from $\mathcal{D}$
		\STATE Build subset $\mathcal{L}$: select the rest samples in $\mathcal{D}$
		\WHILE {$\theta_{k}$ not converged}
		\STATE Update iteration: $k \gets k+1$
		\STATE Select sub-batch $\mathcal{H}_{k}$ of size $0.8m$ from subset $\mathcal{H}$ by SRS
		\STATE Select sub-batch $\mathcal{L}_{k}$ of size $0.2m$ from subset $\mathcal{L}$ by SRS
		\STATE Get batch: $\batch \gets \mathcal{H}_{k}+\mathcal{L}_{k}$
		\STATE Compute gradient : $\nabla J_\batch (\theta_k)=\sum_{i\in \batch}\nabla {{J}_{i}}(\theta)/m$
		\STATE Apply update: $\theta_{k+1}=\theta_{k}-\eta*\nabla J_\batch (\theta_k)$    
		\ENDWHILE
	\end{algorithmic} 
	\caption{Typical batch SGD: t-SNE embedding}  
	\label{typical batch SGD tsne}
\end{algorithm}

\subsection{Experiments on Synthetic Dataset}
To evaluate the proposed sampling scheme, we start with experiments on synthetic PWL-Curve dataset built in \cite{shalev2013stochastic}. The dataset contains 50000 one-dimensional piece-wise linear curves and the problem we consider is encoding the values of curve $f$ into key parameters (e.g. bias, slope) so that we can rebuild $f$ from it. We compare the performance of typical batch SGD using t-SNE (TBS+t-SNE) against conventional Minibatch SGD by measuring average loss value on training set and validation set. We also combine typicality sampling scheme with Adam (Adam+TS) to test the versatility of the proposed batch selection method. The model we use is a one layer Convolutional Neural Network (CNN) with the kernel $[1,-2,1]$. Trainings are conducted using fixed stepsize without any specific tuning skill.

\begin{figure*}[!t]
	\centering
	\begin{subfigure}[t]{0.32\textwidth}
		\includegraphics[width=\textwidth]{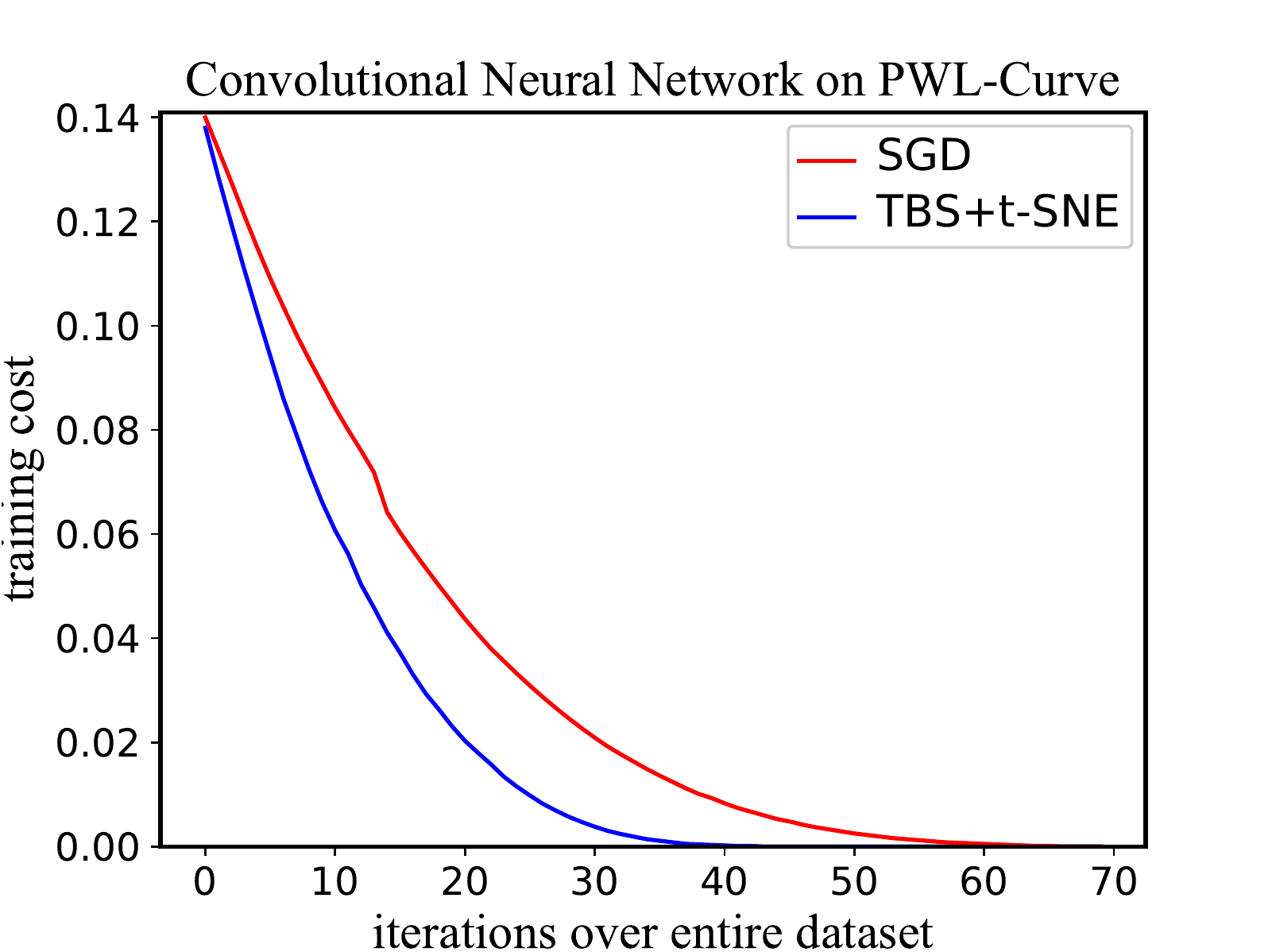}
	\end{subfigure}
	\hfill
	\begin{subfigure}[t]{0.32\textwidth}
		\includegraphics[width=\textwidth]{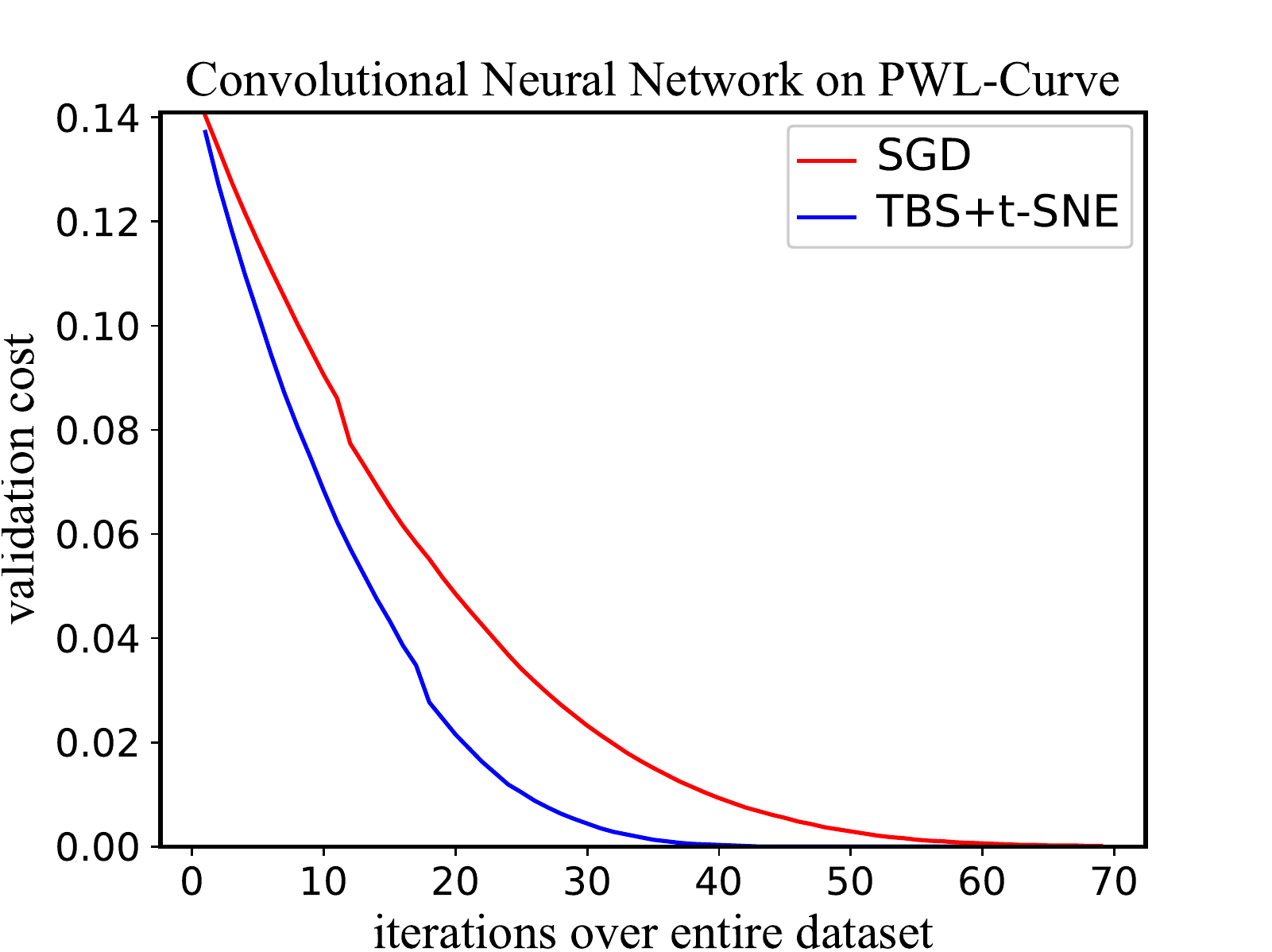}
	\end{subfigure}
	\hfill
	\begin{subfigure}[t]{0.32\textwidth}
		\includegraphics[width=\textwidth]{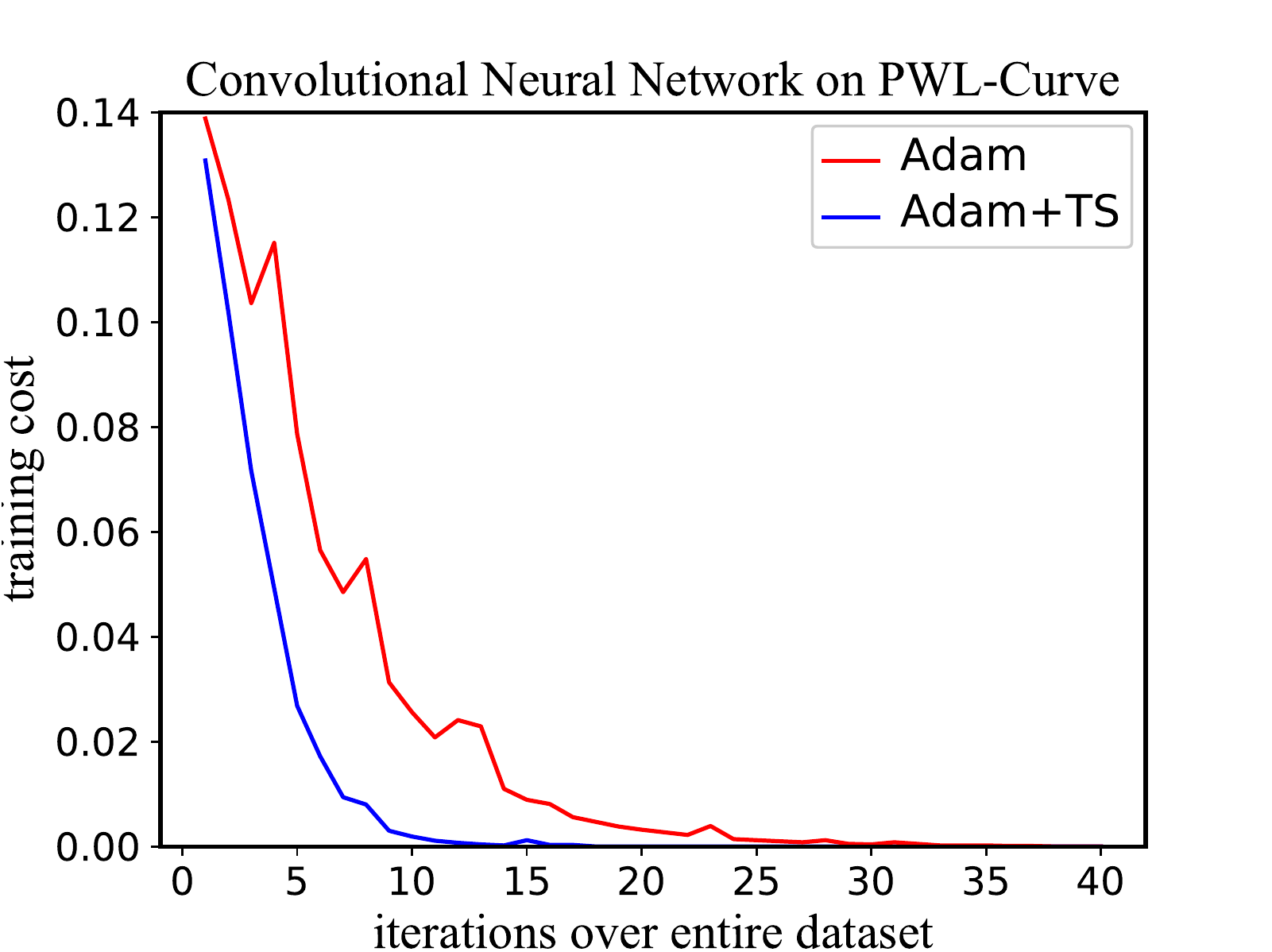}
	\end{subfigure} 
	\caption{Experiments on synthetic dataset. The left and middle columns present the curves of training loss and validation loss against Minibatch SGD, the right column presents the curves of training cost against Adam.}
	\label{fig:synthetic}
\end{figure*}

\begin{figure*}[!t]
	\centering
	\begin{subfigure}[t]{0.45\textwidth}
		\captionsetup[subfigure]{singlelinecheck=off,justification=raggedright}
		\begin{subfigure}[t]{\textwidth}
			\setlength{\abovecaptionskip}{-0.1cm}
			\setlength{\belowcaptionskip}{0.2cm}
			\subcaption*{~~Example a}
			\includegraphics[width=0.30\textwidth]{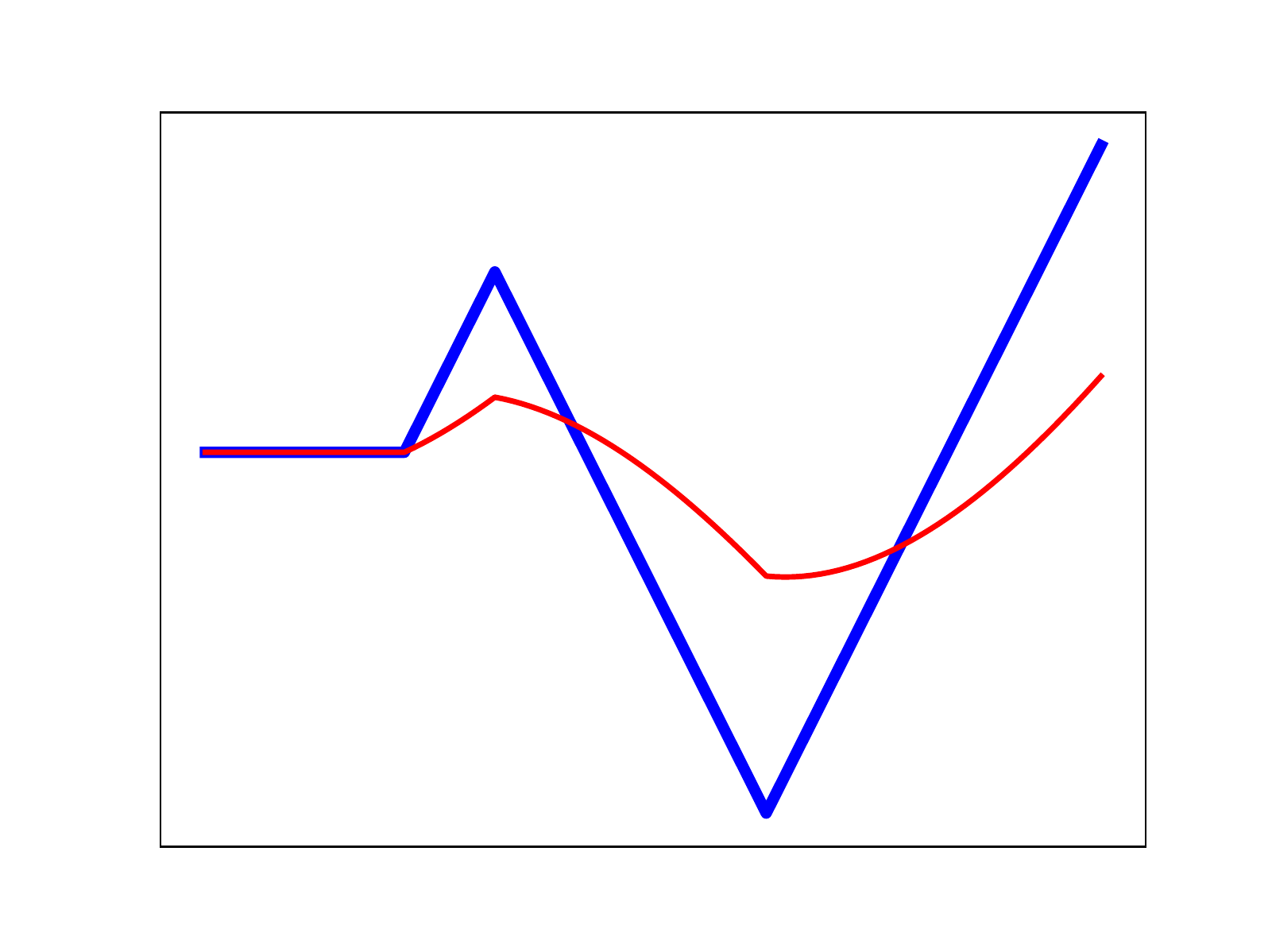}
			\includegraphics[width=0.30\textwidth]{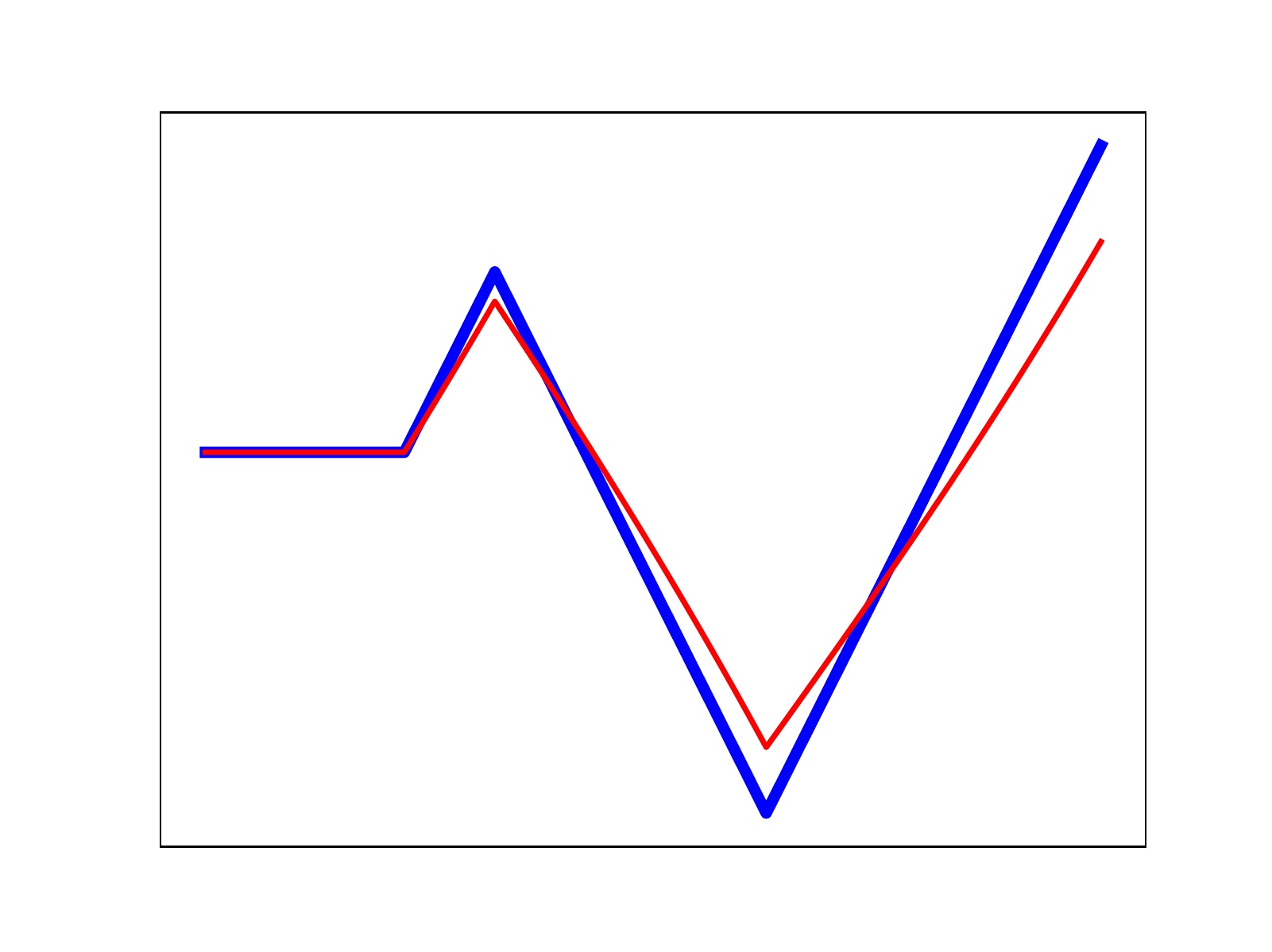}
			\includegraphics[width=0.30\textwidth]{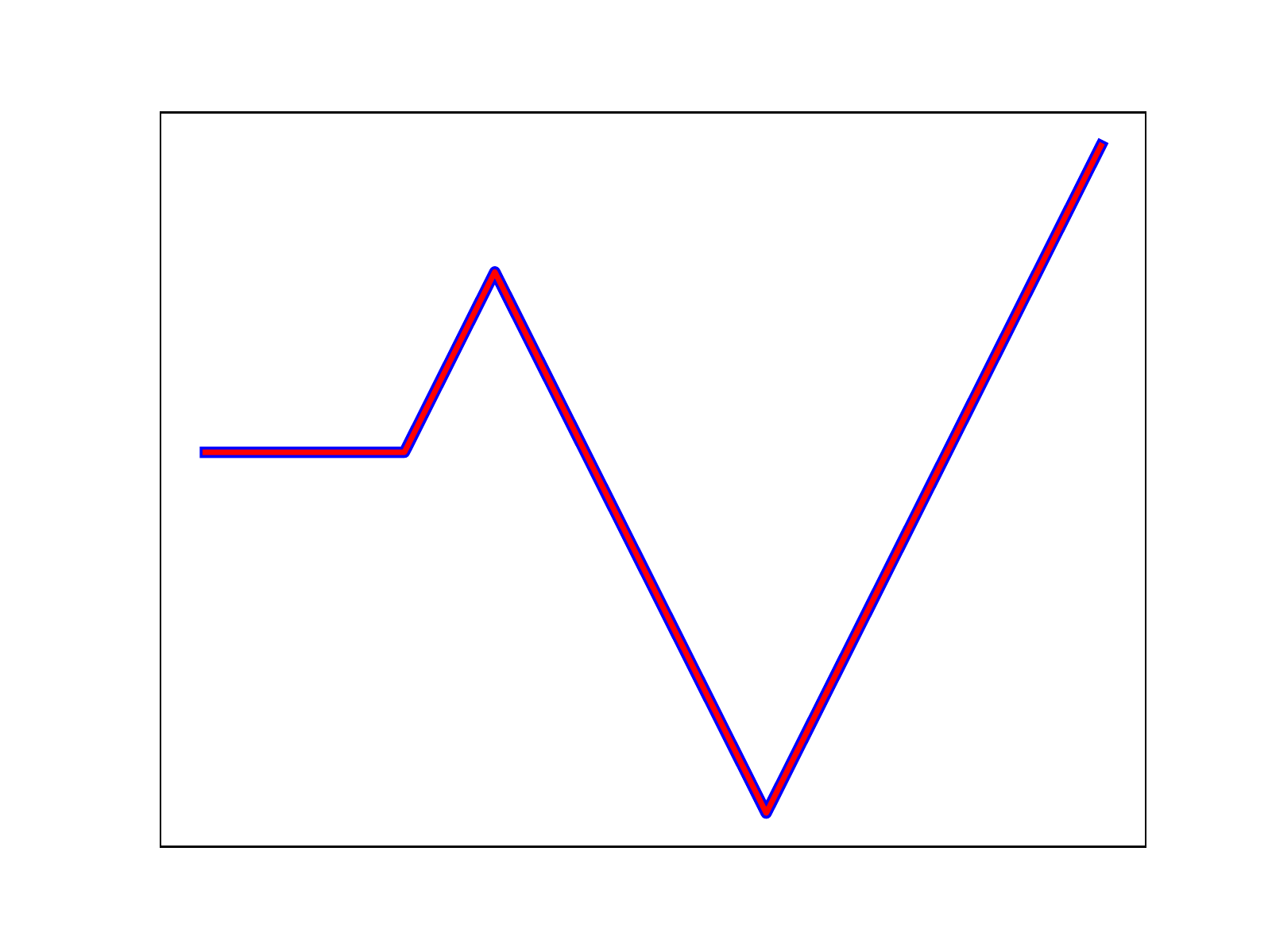}	
		\end{subfigure}
		\begin{subfigure}[t]{\textwidth}
			\setlength{\abovecaptionskip}{-0.1cm}
			\setlength{\belowcaptionskip}{0.2cm}
			\subcaption*{~~Example b}
			\includegraphics[width=0.30\textwidth]{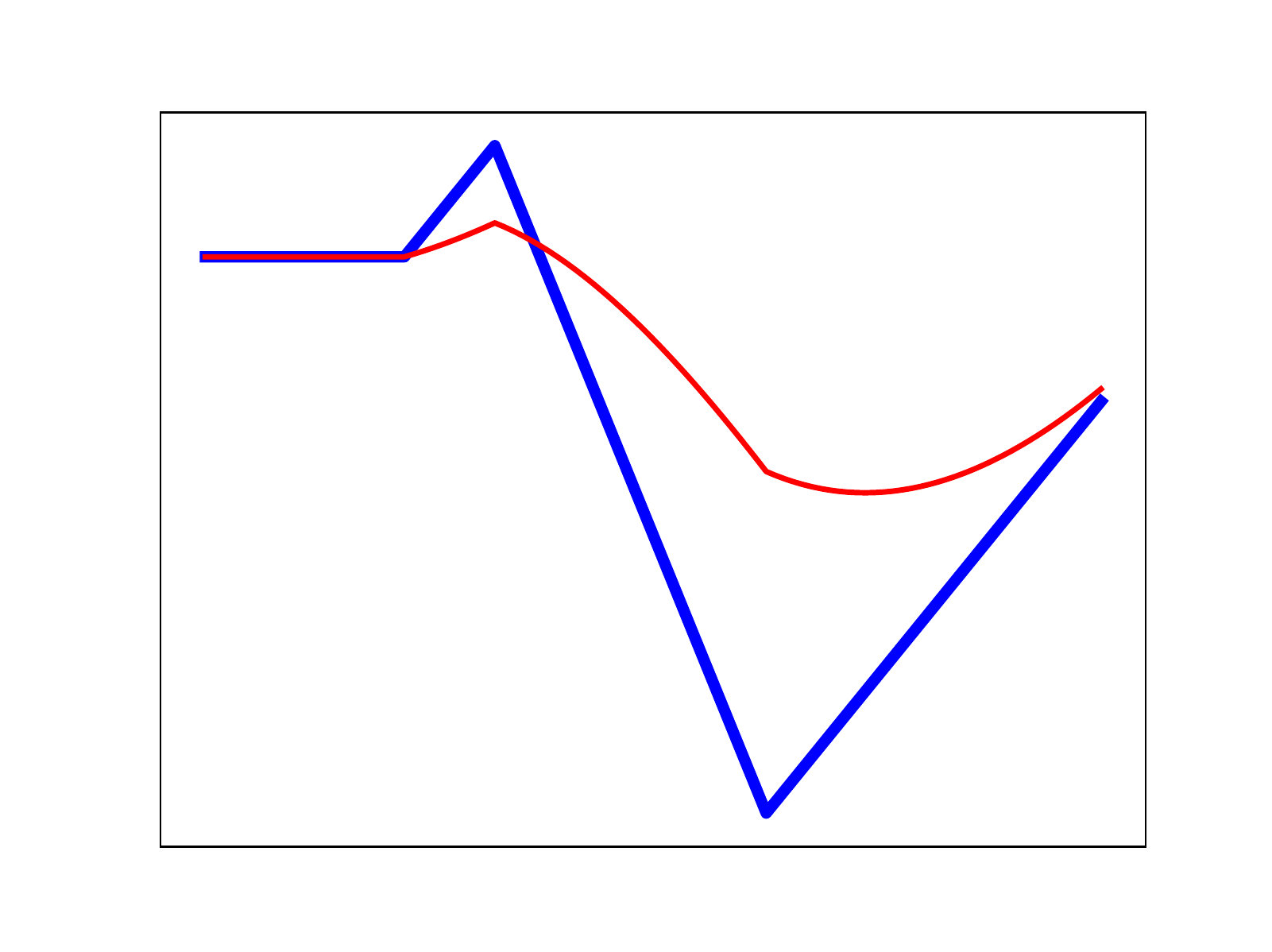}
			\includegraphics[width=0.30\textwidth]{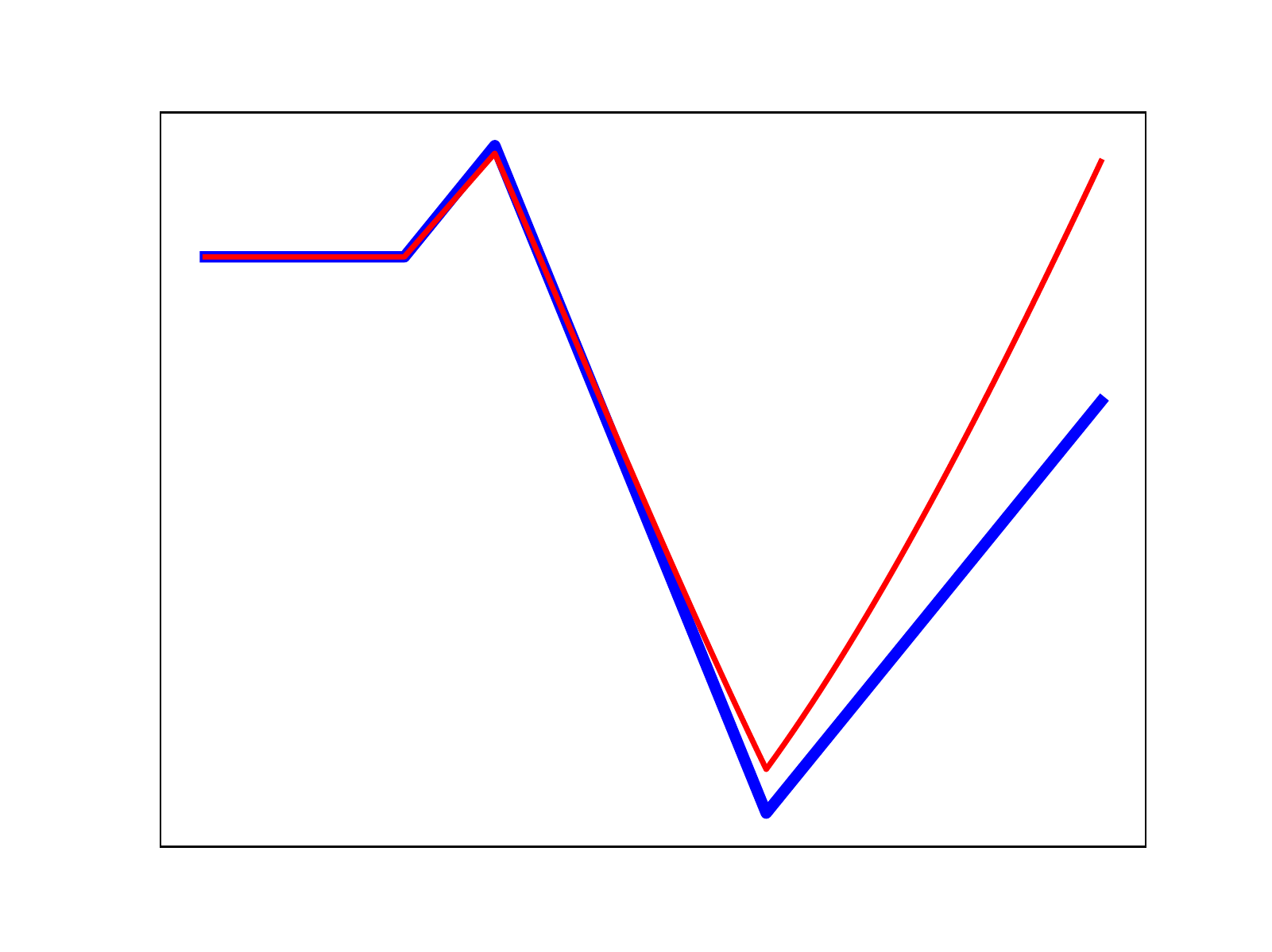}
			\includegraphics[width=0.30\textwidth]{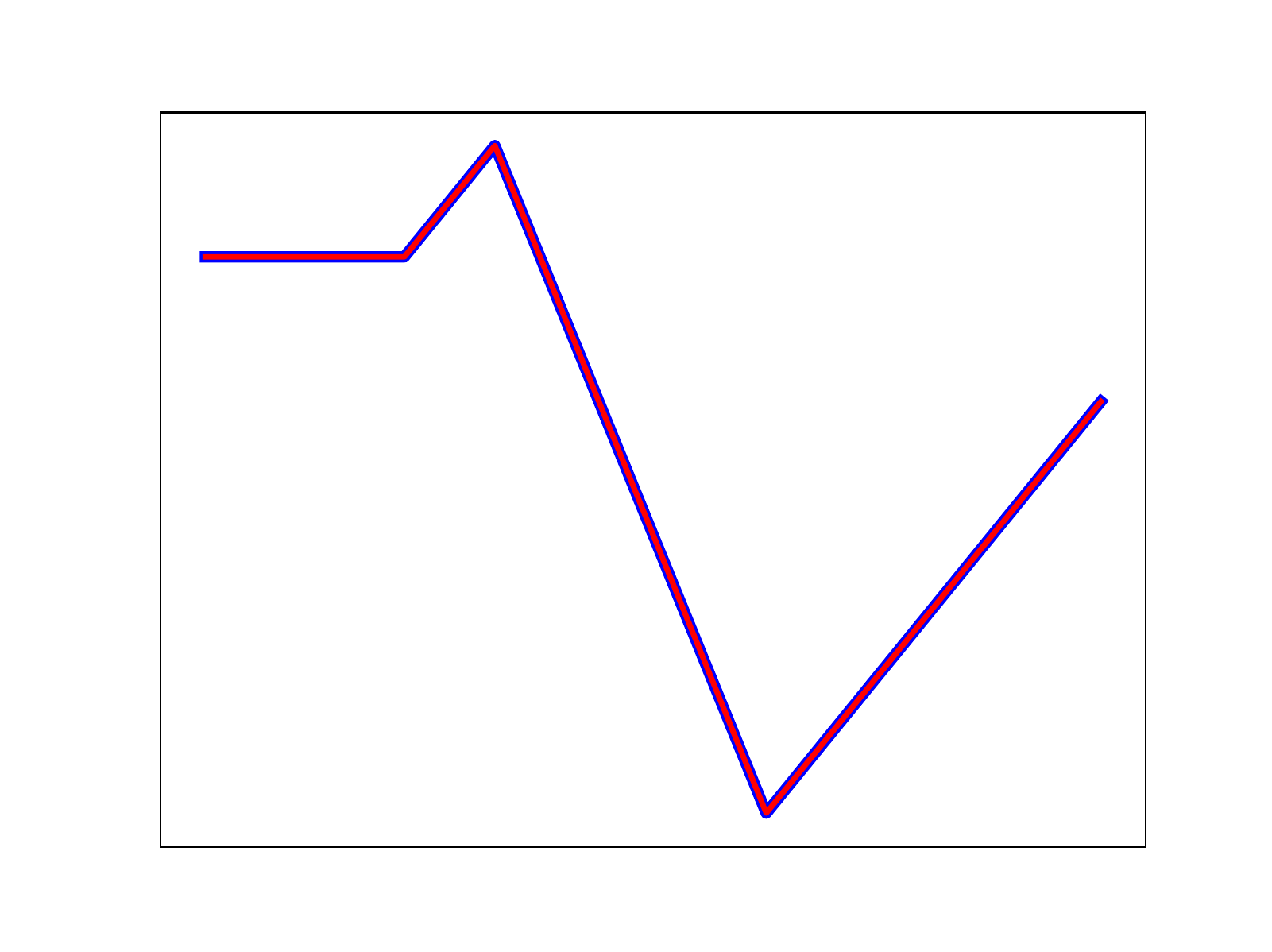}  
		\end{subfigure}
		\begin{subfigure}[t]{\textwidth}
			\setlength{\abovecaptionskip}{-0.1cm}
			\setlength{\belowcaptionskip}{0.2cm}
			\subcaption*{~~Example c}
			\includegraphics[width=0.30\textwidth]{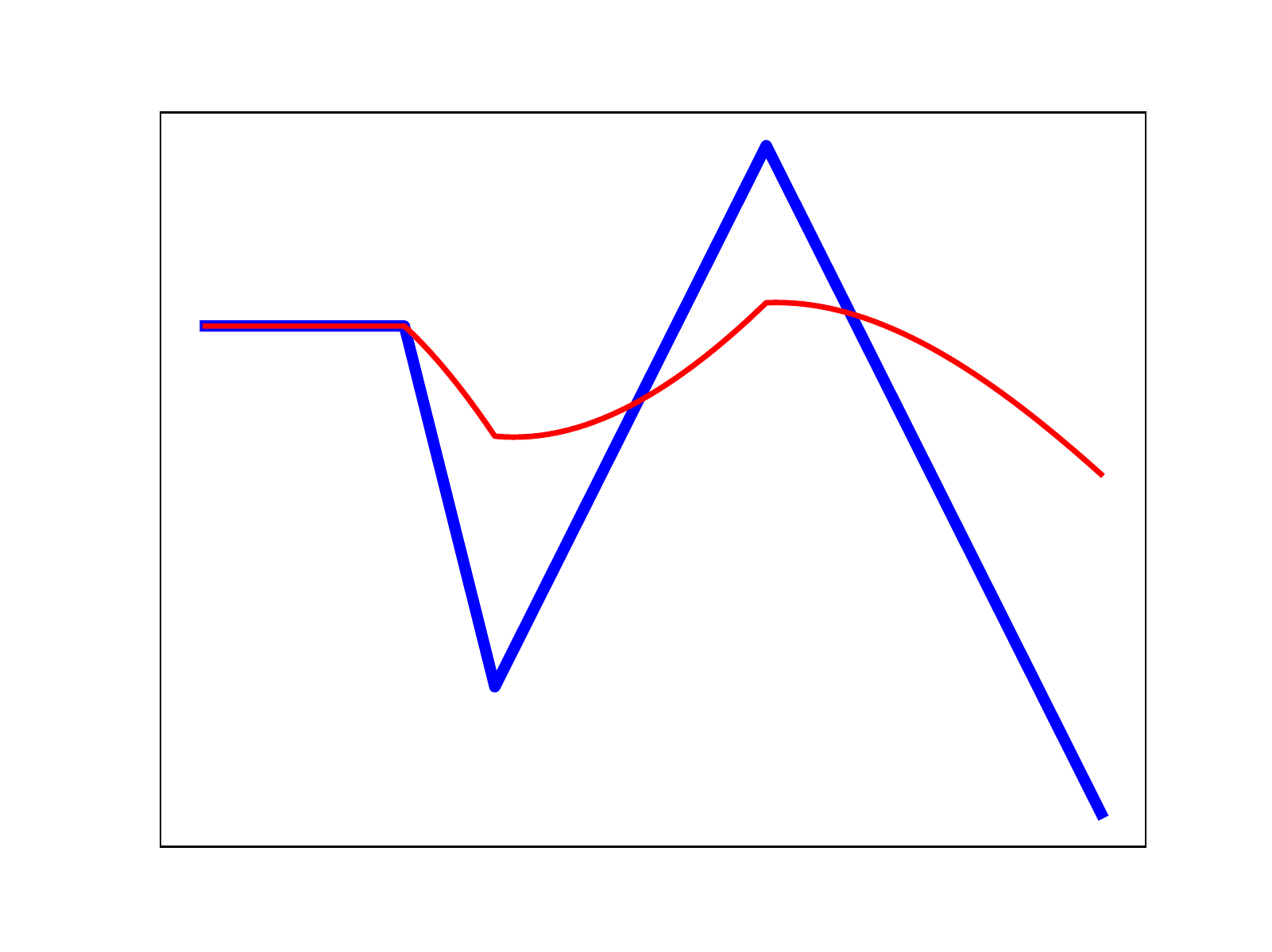}
			\includegraphics[width=0.30\textwidth]{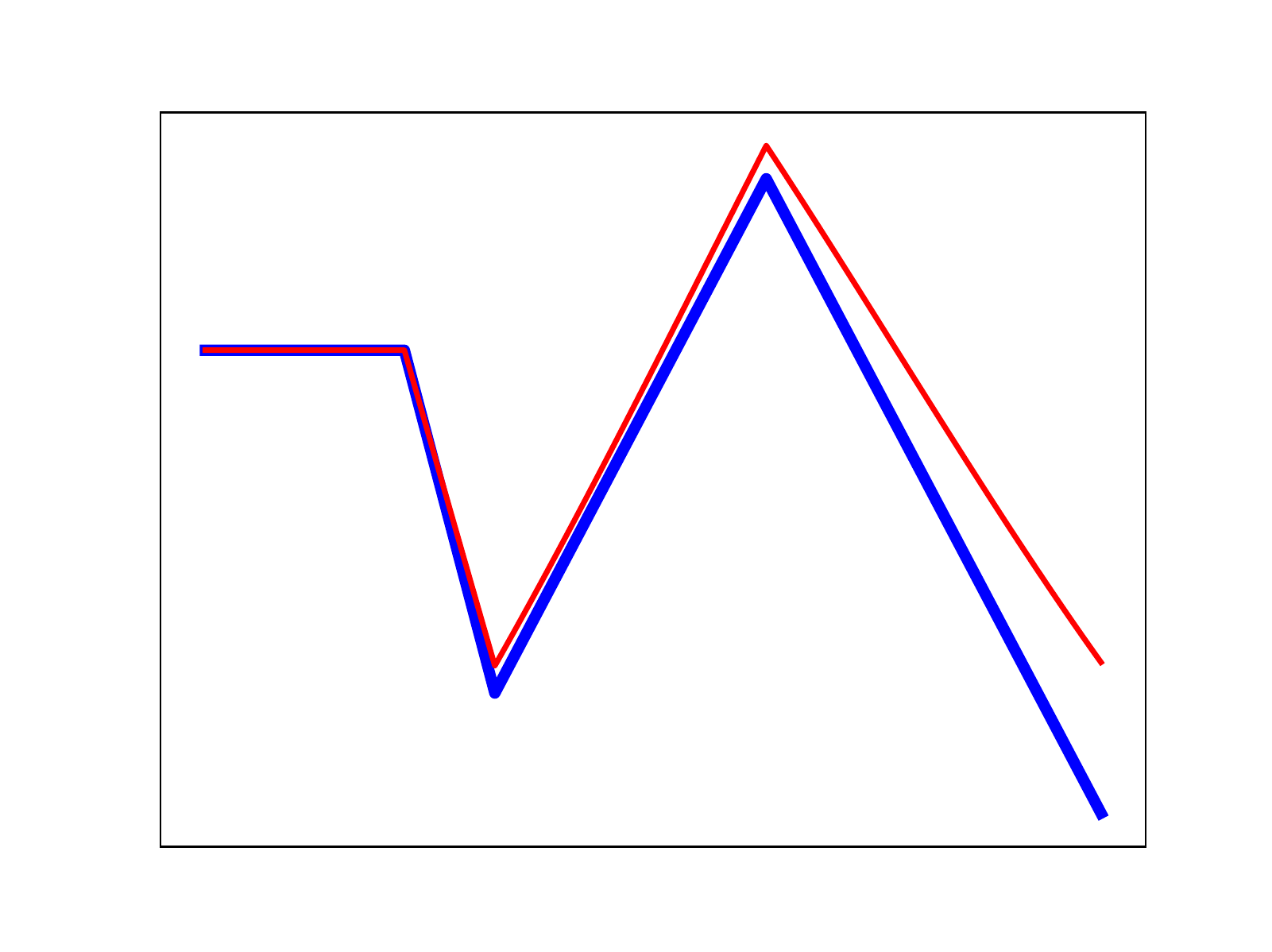}
			\includegraphics[width=0.30\textwidth]{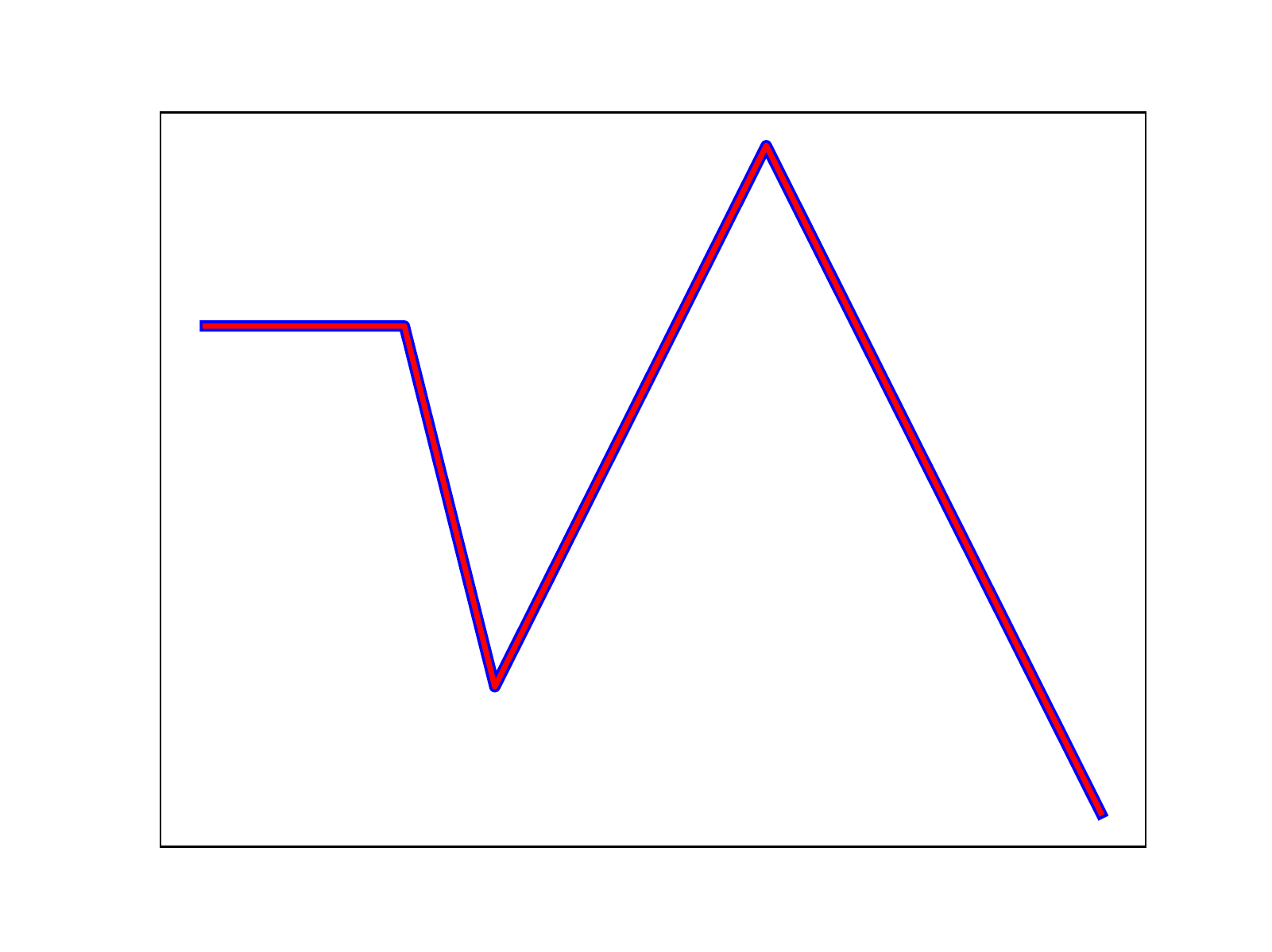}
		\end{subfigure}
		\begin{subfigure}[t]{\textwidth}
			\setlength{\abovecaptionskip}{-0.1cm}
			\setlength{\belowcaptionskip}{0.2cm}
			\subcaption*{~~Example d}
			\includegraphics[width=0.30\textwidth]{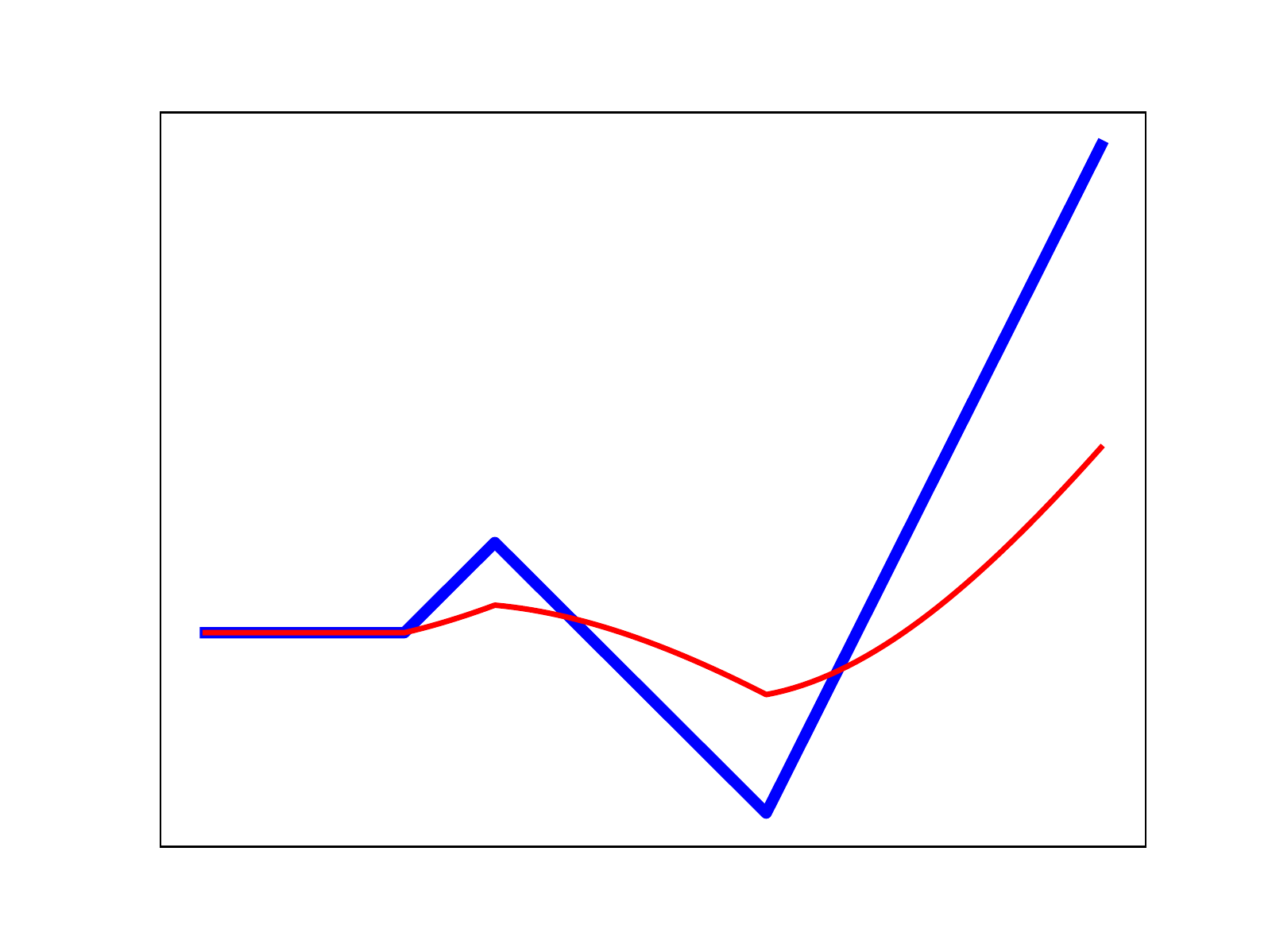}
			\includegraphics[width=0.30\textwidth]{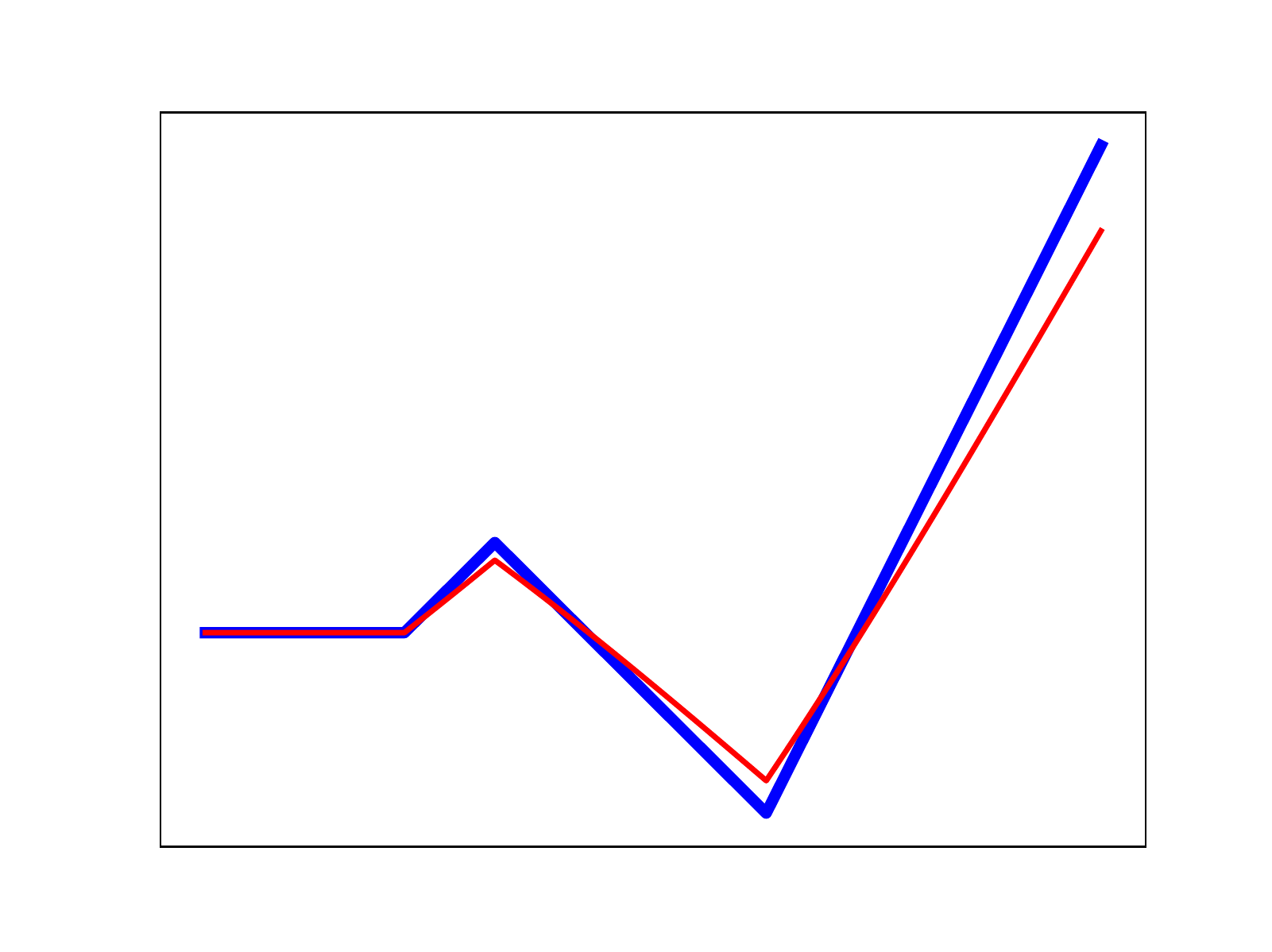}
			\includegraphics[width=0.30\textwidth]{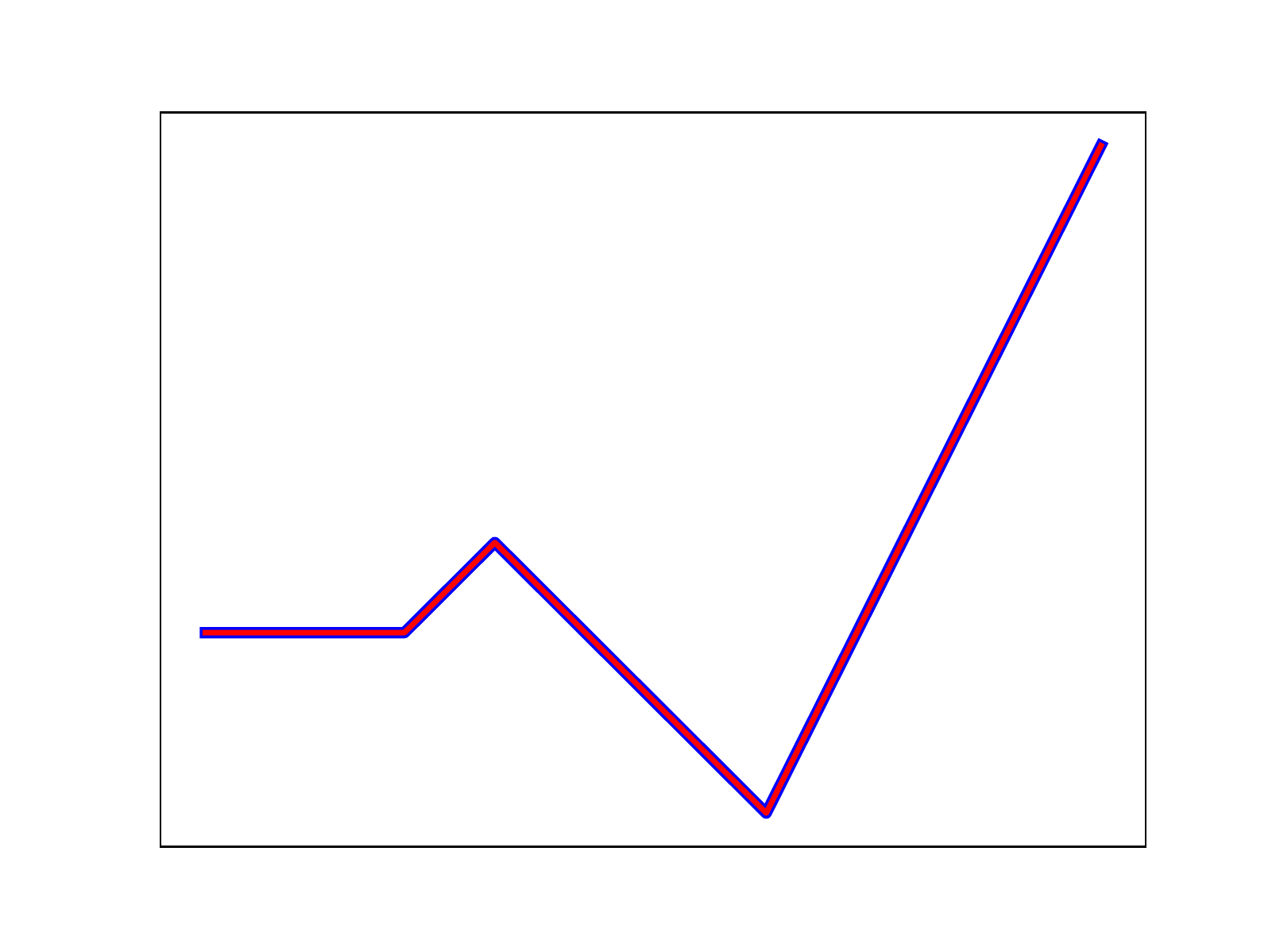}
		\end{subfigure}\vspace{0.3cm}
		\begin{subfigure}[t]{0.9\textwidth}
			\centering
			~~\includegraphics[width=\textwidth]{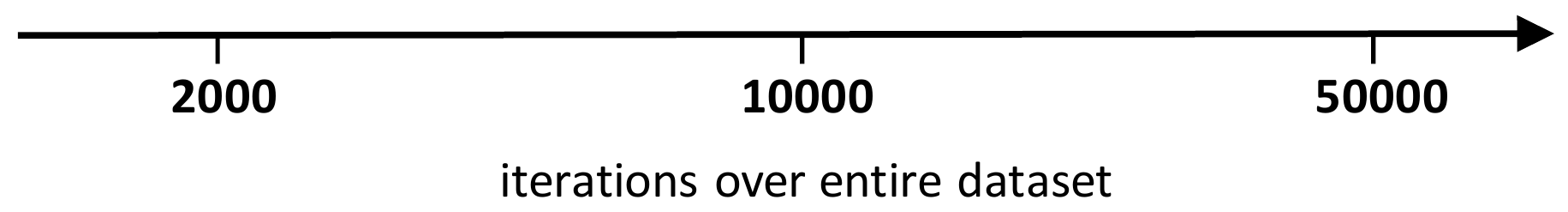}
		\end{subfigure}
		\caption{Mini-Batch SGD}
	\end{subfigure}
	\begin{subfigure}[t]{0.45\textwidth}
		\captionsetup[subfigure]{singlelinecheck=off,justification=raggedright}
		\begin{subfigure}[t]{\textwidth}
			\setlength{\abovecaptionskip}{-0.1cm}
			\setlength{\belowcaptionskip}{0.2cm}
			\subcaption*{~~Example a}
			\includegraphics[width=0.30\textwidth]{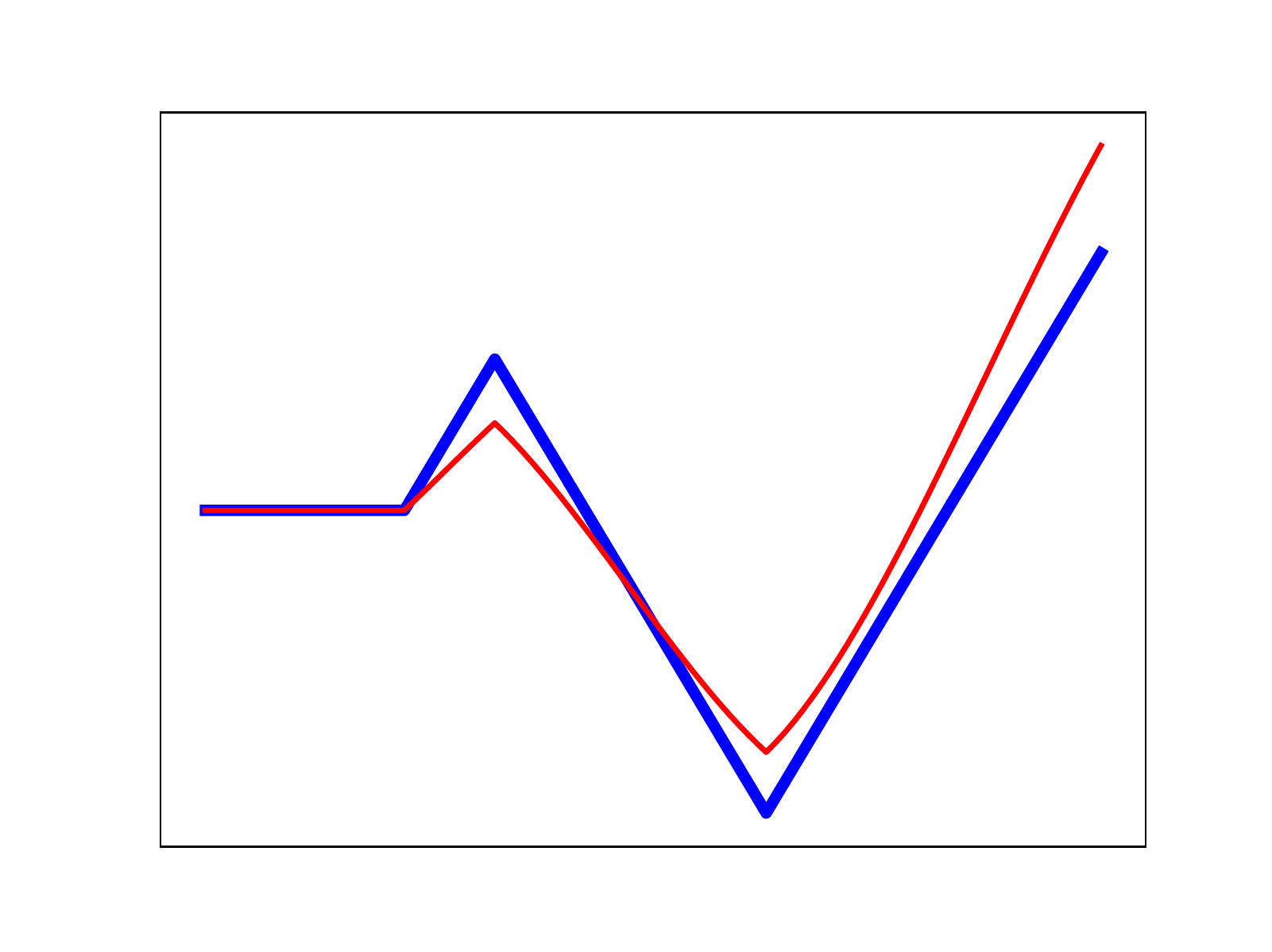}
			\includegraphics[width=0.30\textwidth]{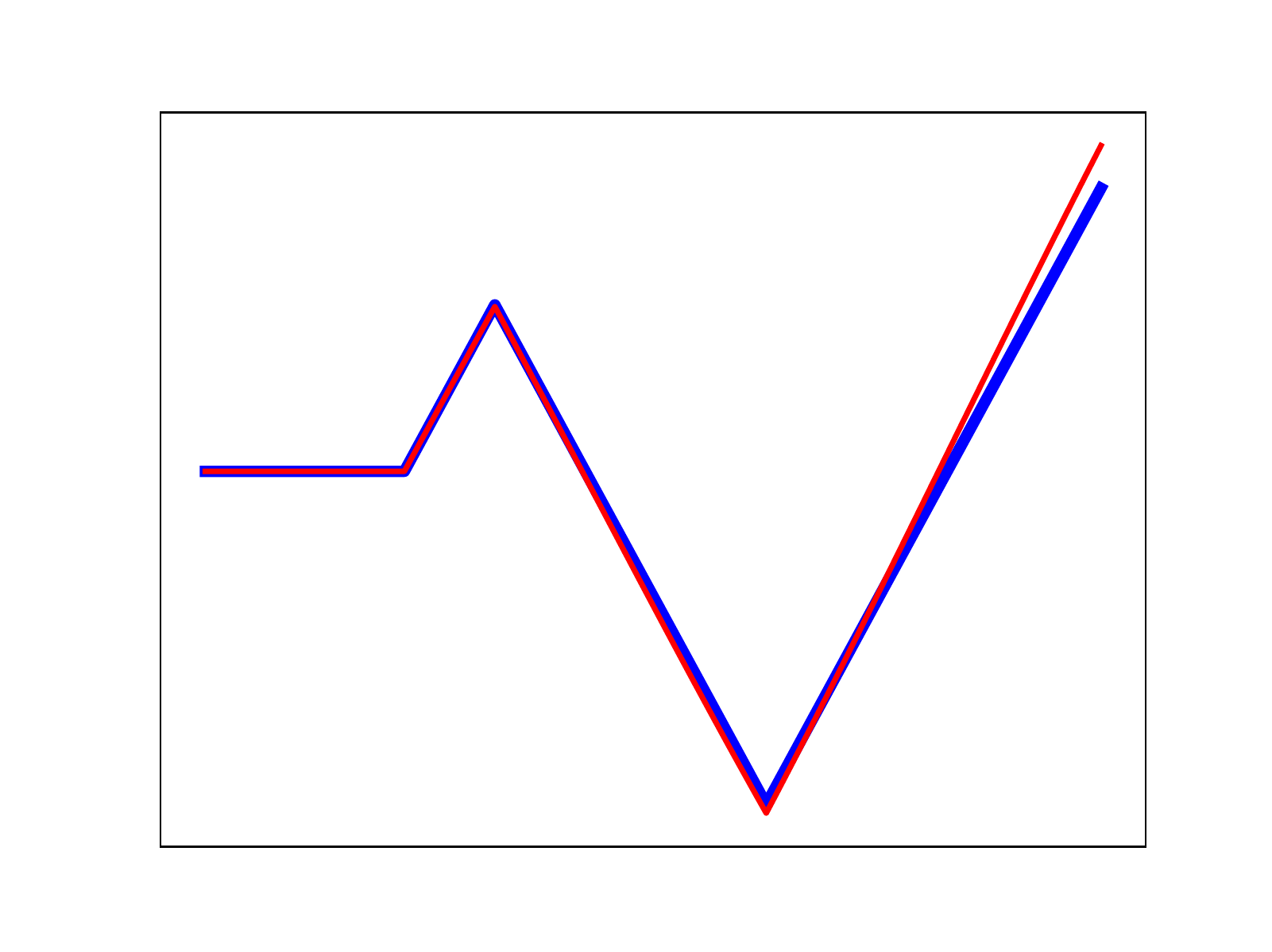}
			\includegraphics[width=0.30\textwidth]{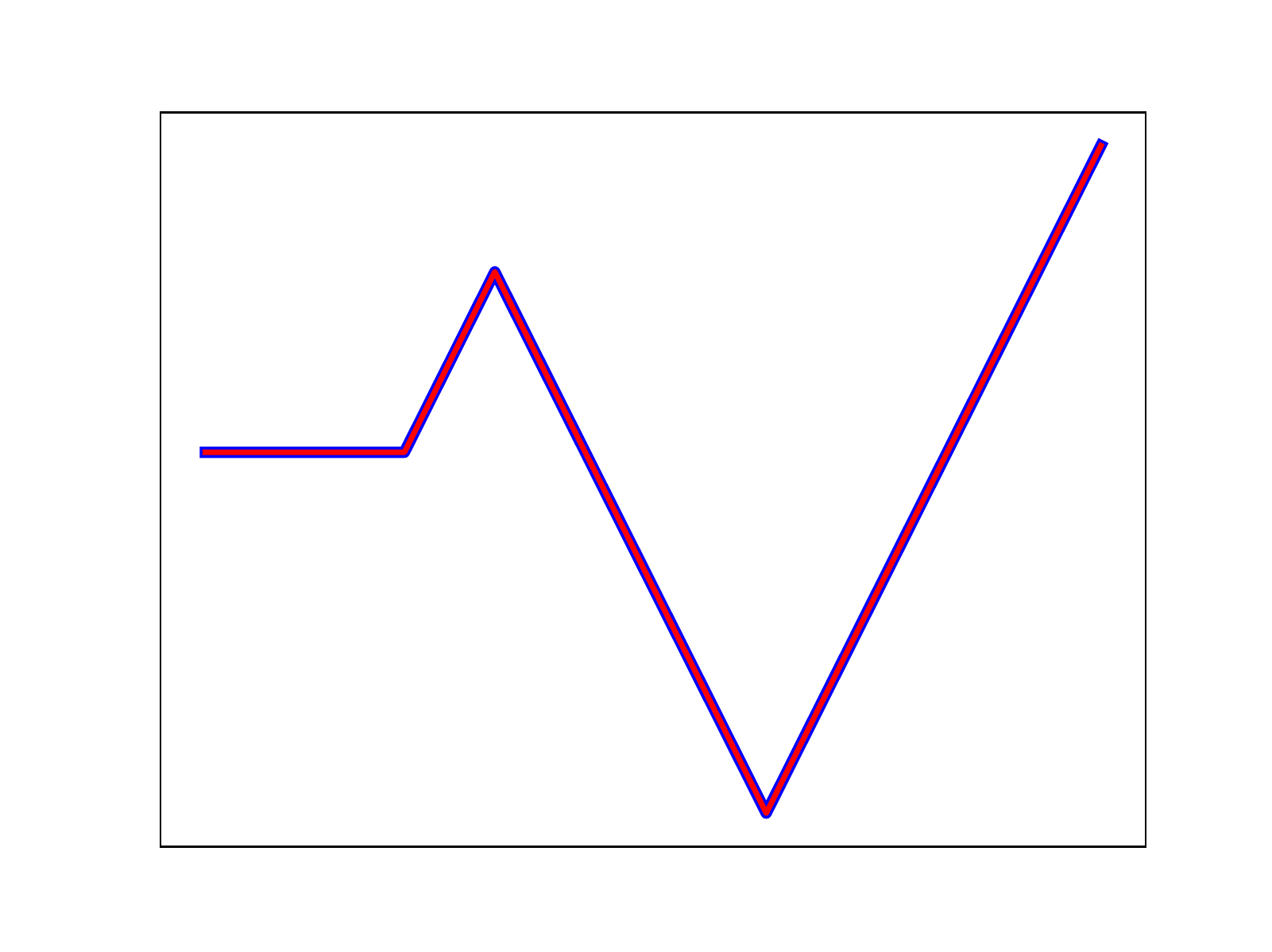}
		\end{subfigure}
		\begin{subfigure}[t]{\textwidth}
			\setlength{\abovecaptionskip}{-0.1cm}
			\setlength{\belowcaptionskip}{0.2cm}
			\subcaption*{~~Example b}
			\includegraphics[width=0.30\textwidth]{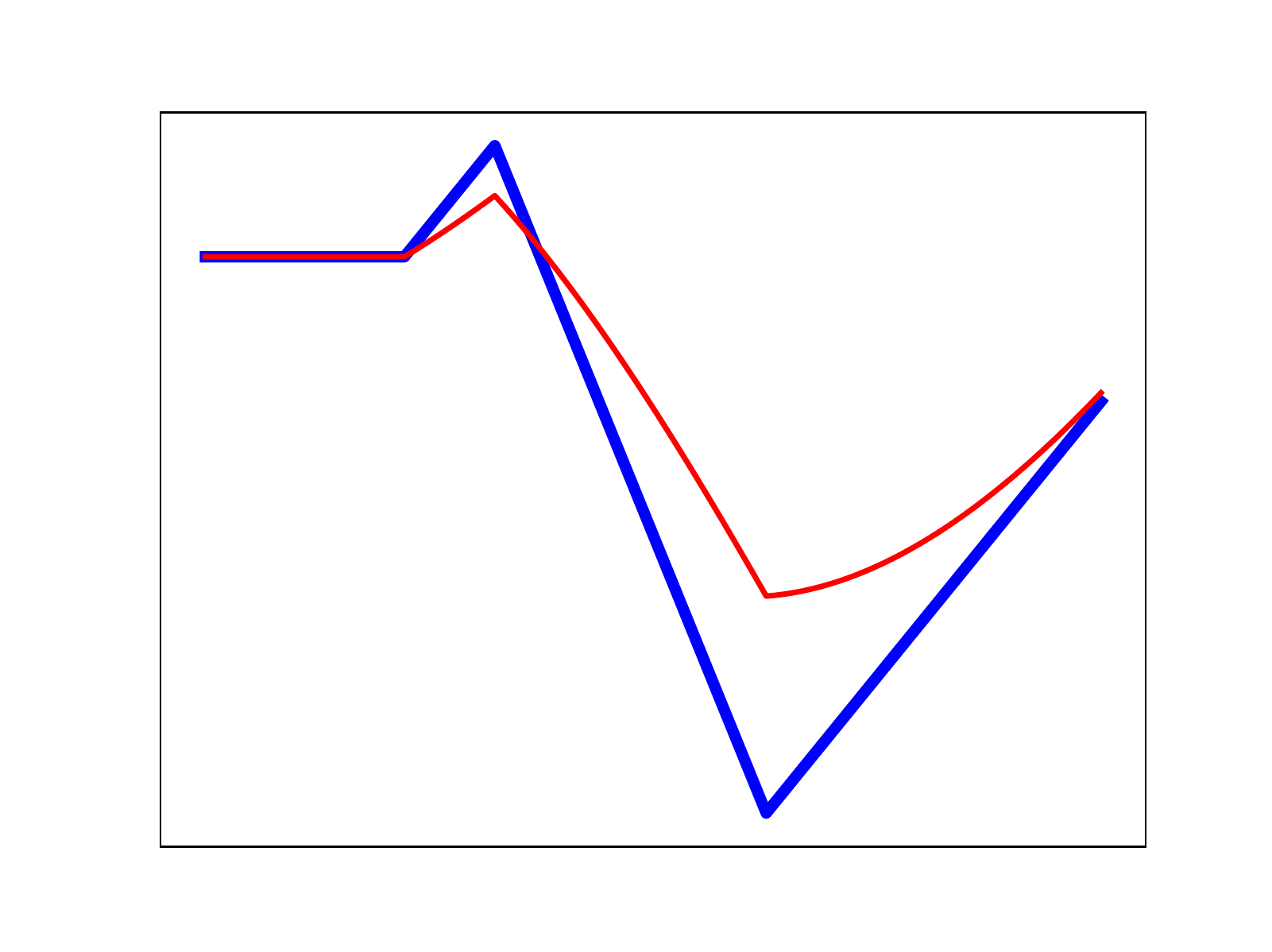}
			\includegraphics[width=0.30\textwidth]{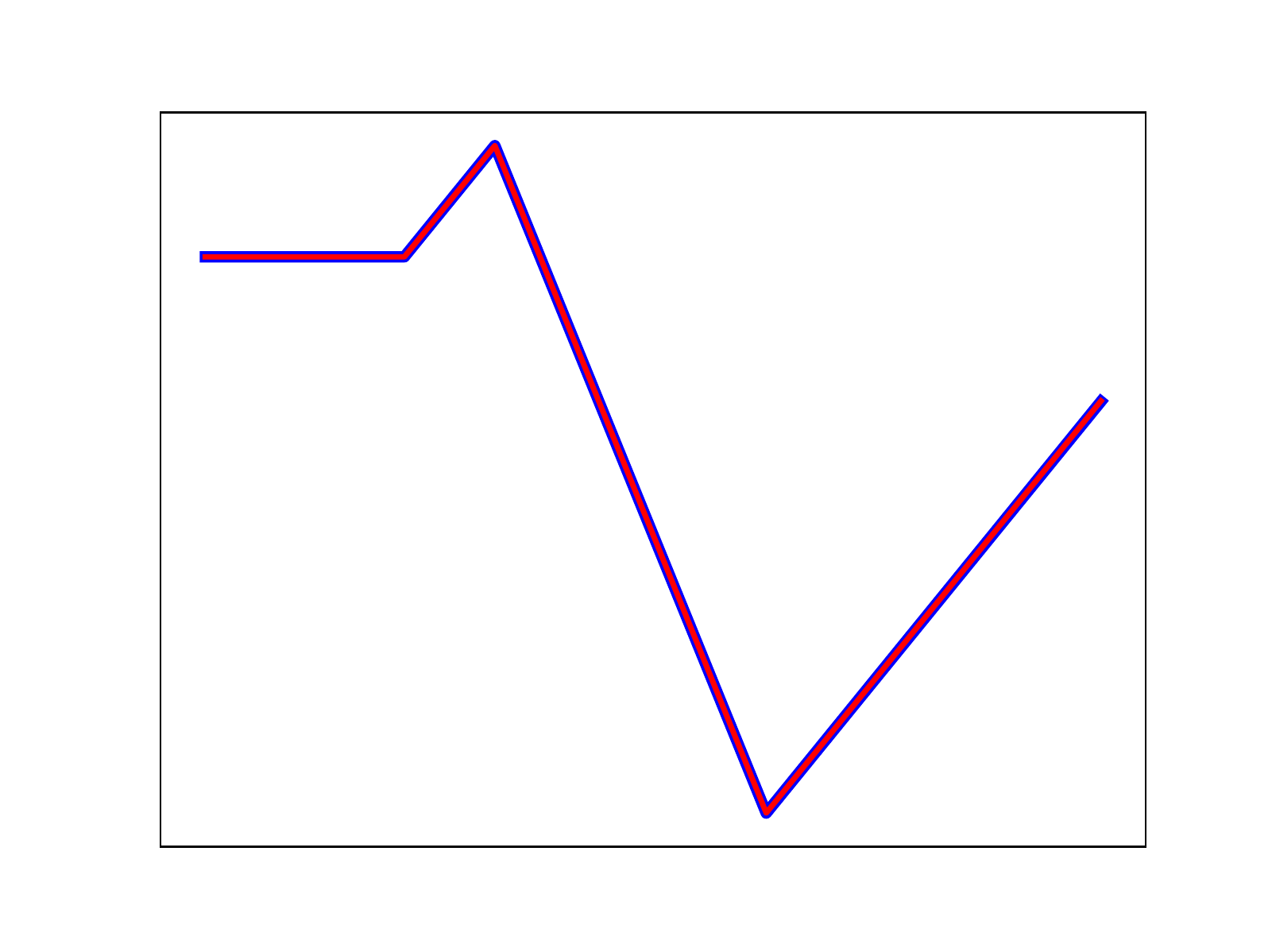}
			\includegraphics[width=0.30\textwidth]{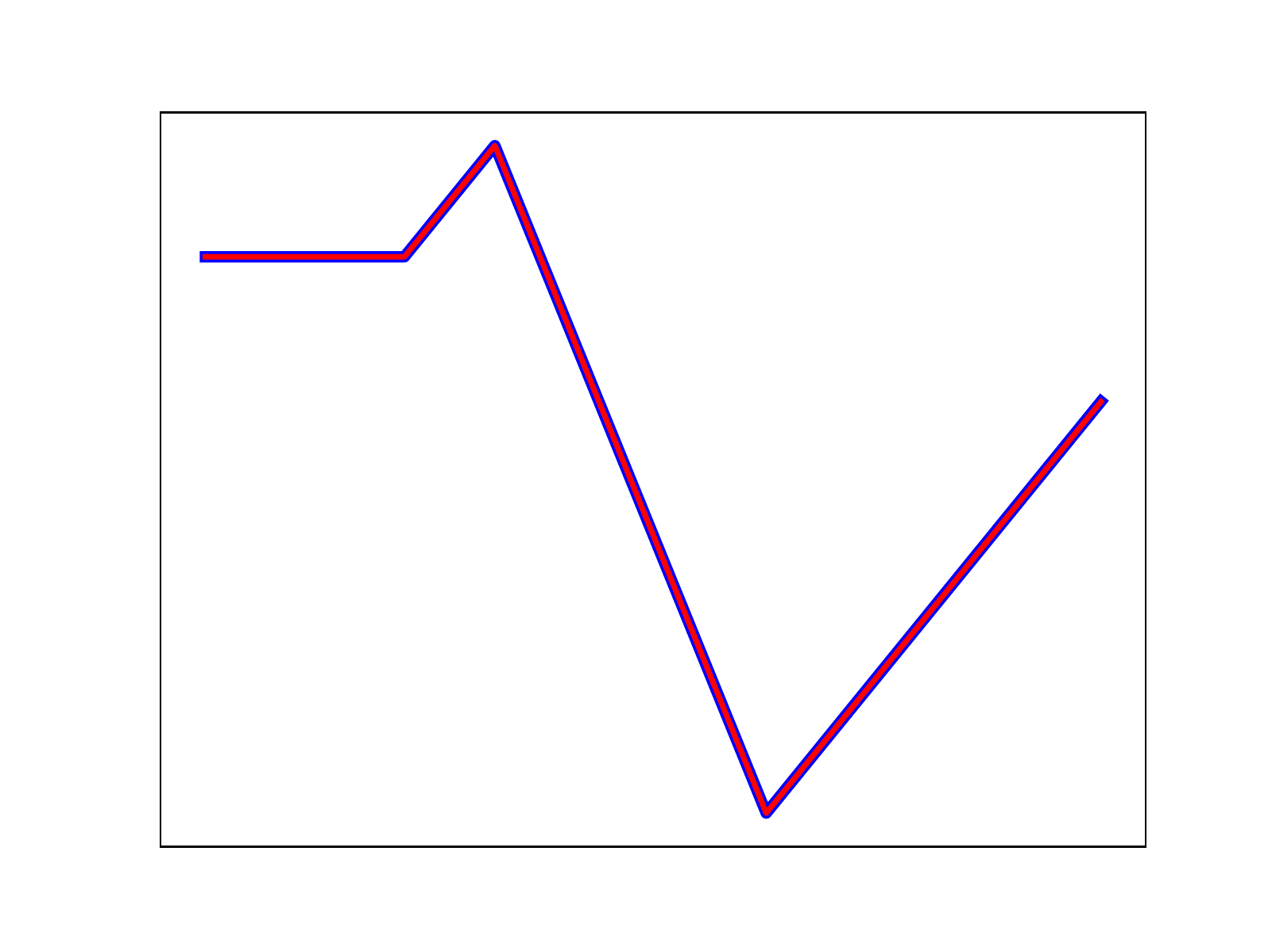}
		\end{subfigure}
		\begin{subfigure}[t]{\textwidth}
			\setlength{\abovecaptionskip}{-0.1cm}
			\setlength{\belowcaptionskip}{0.2cm}
			\subcaption*{~~Example c}
			\includegraphics[width=0.30\textwidth]{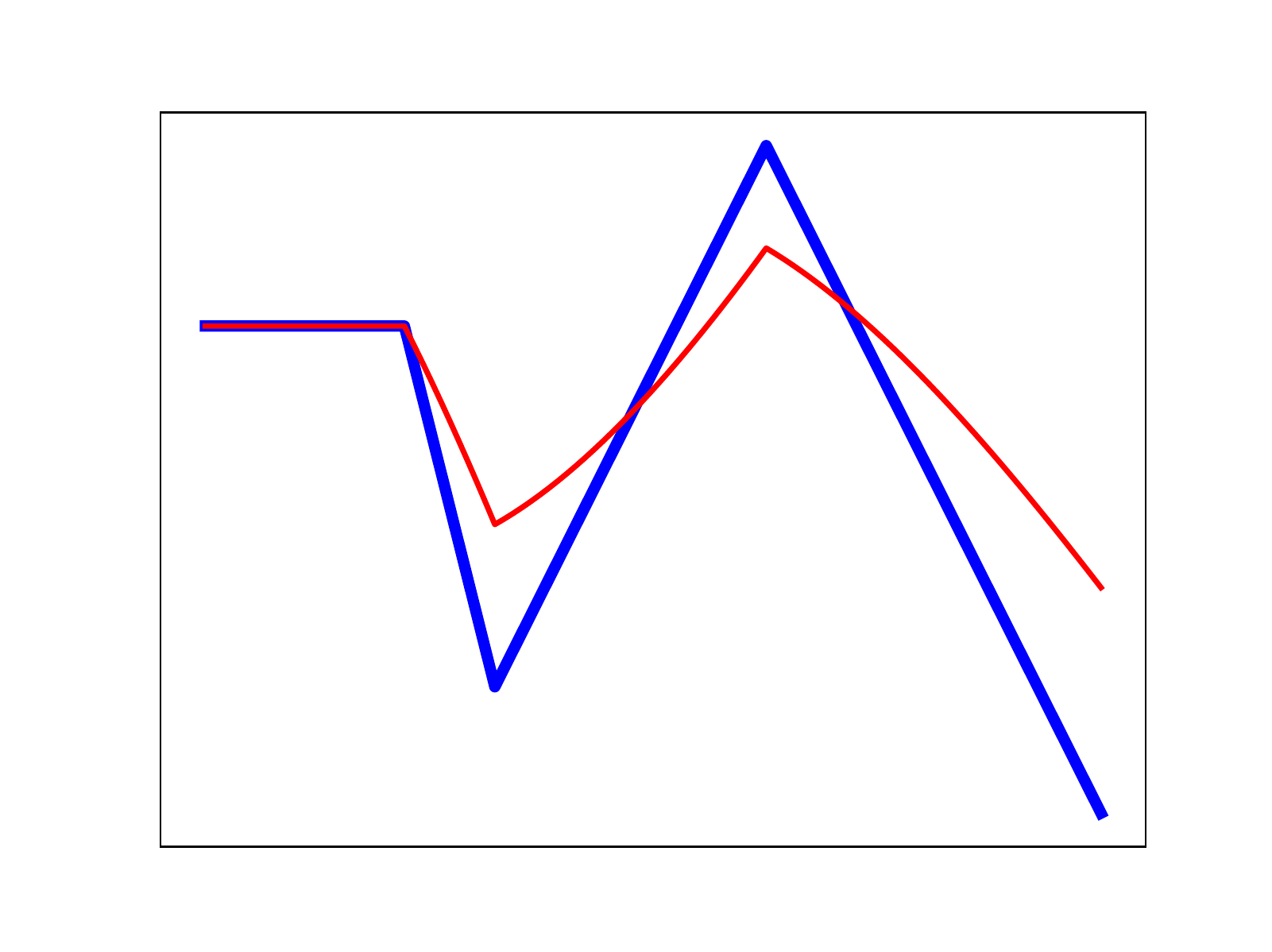}
			\includegraphics[width=0.30\textwidth]{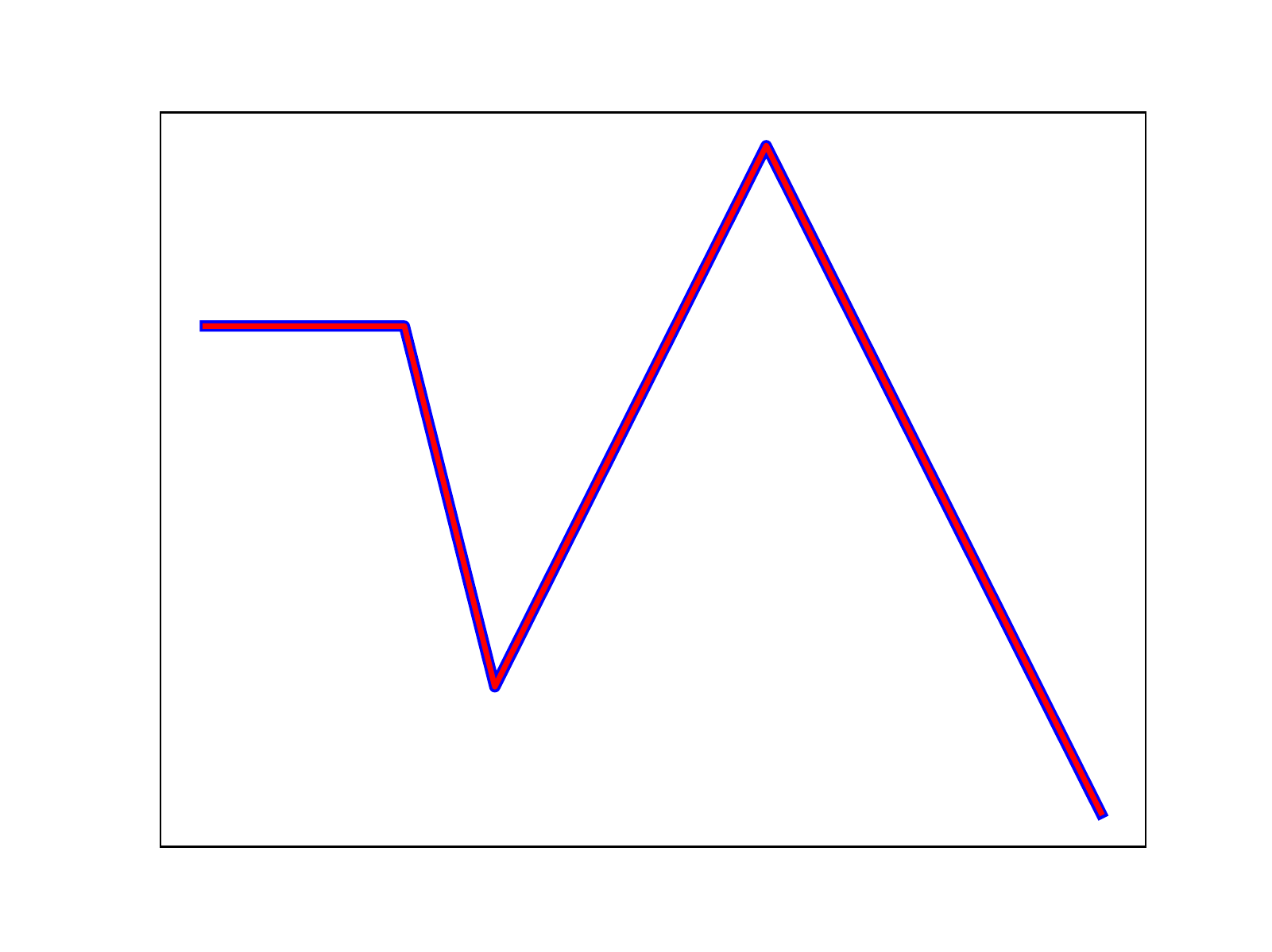}
			\includegraphics[width=0.30\textwidth]{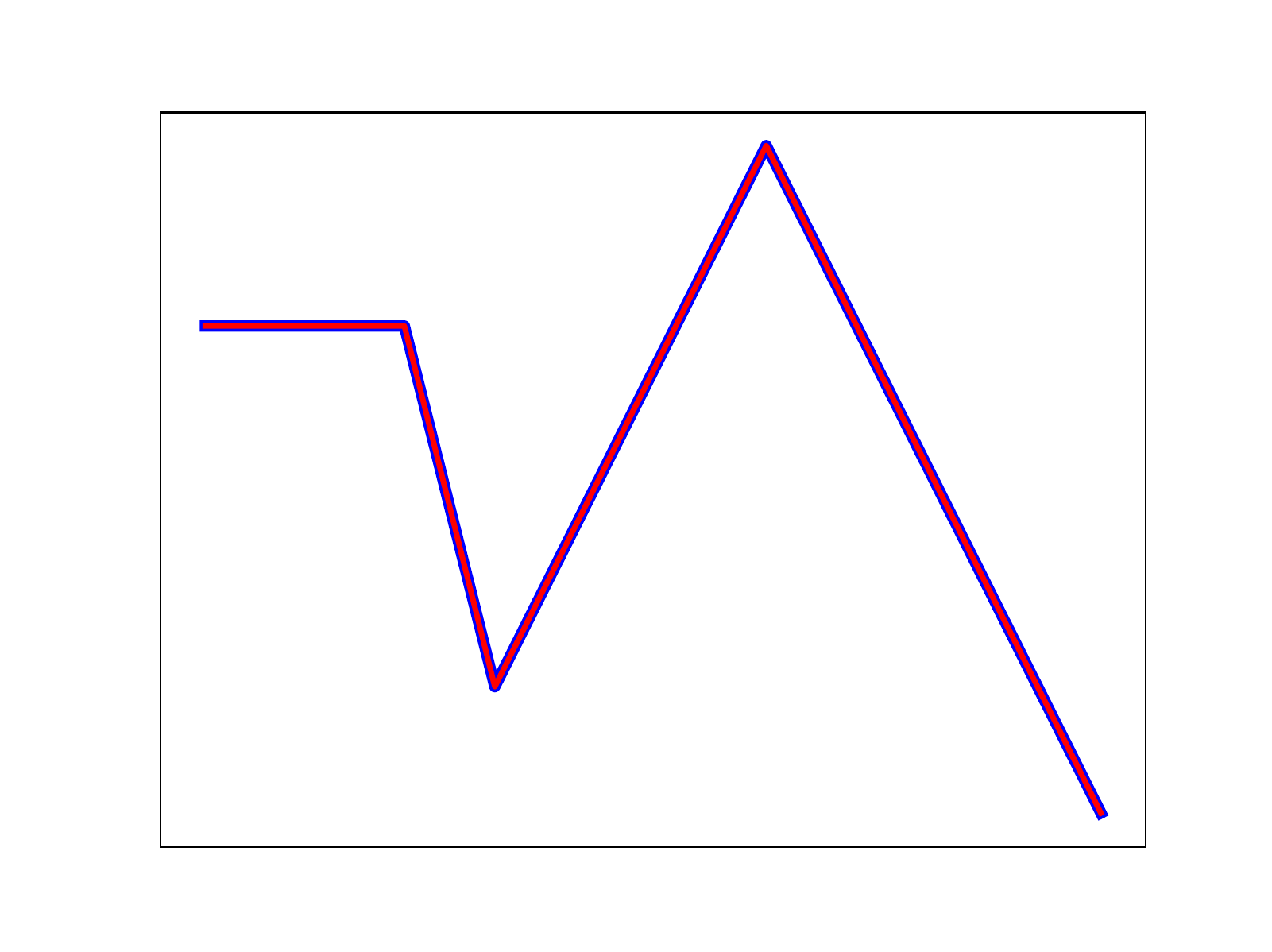}
		\end{subfigure}
		\begin{subfigure}[t]{\textwidth}
			\setlength{\abovecaptionskip}{-0.1cm}
			\setlength{\belowcaptionskip}{0.2cm}
			\subcaption*{~~Example d}
			\includegraphics[width=0.30\textwidth]{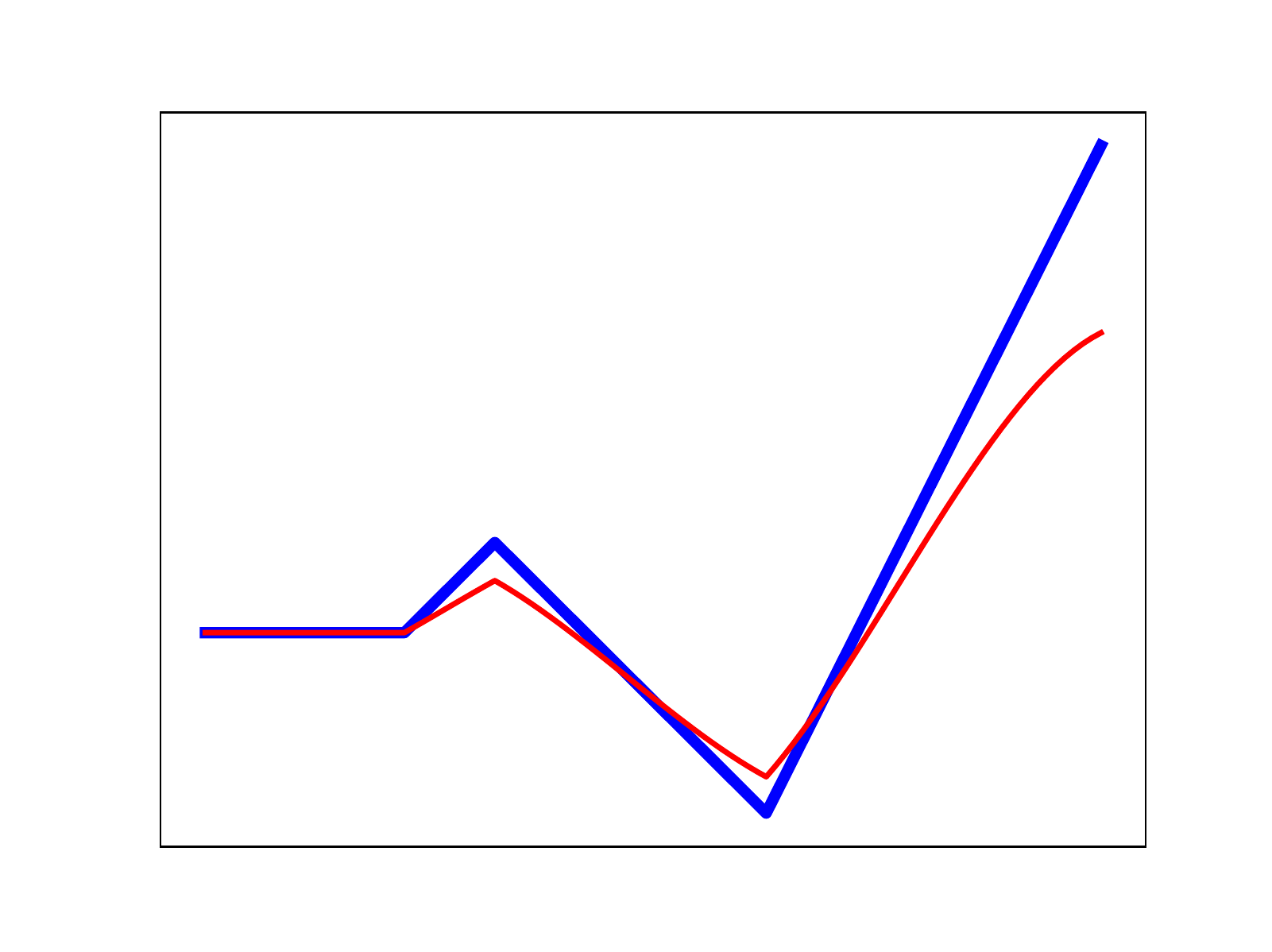}
			\includegraphics[width=0.30\textwidth]{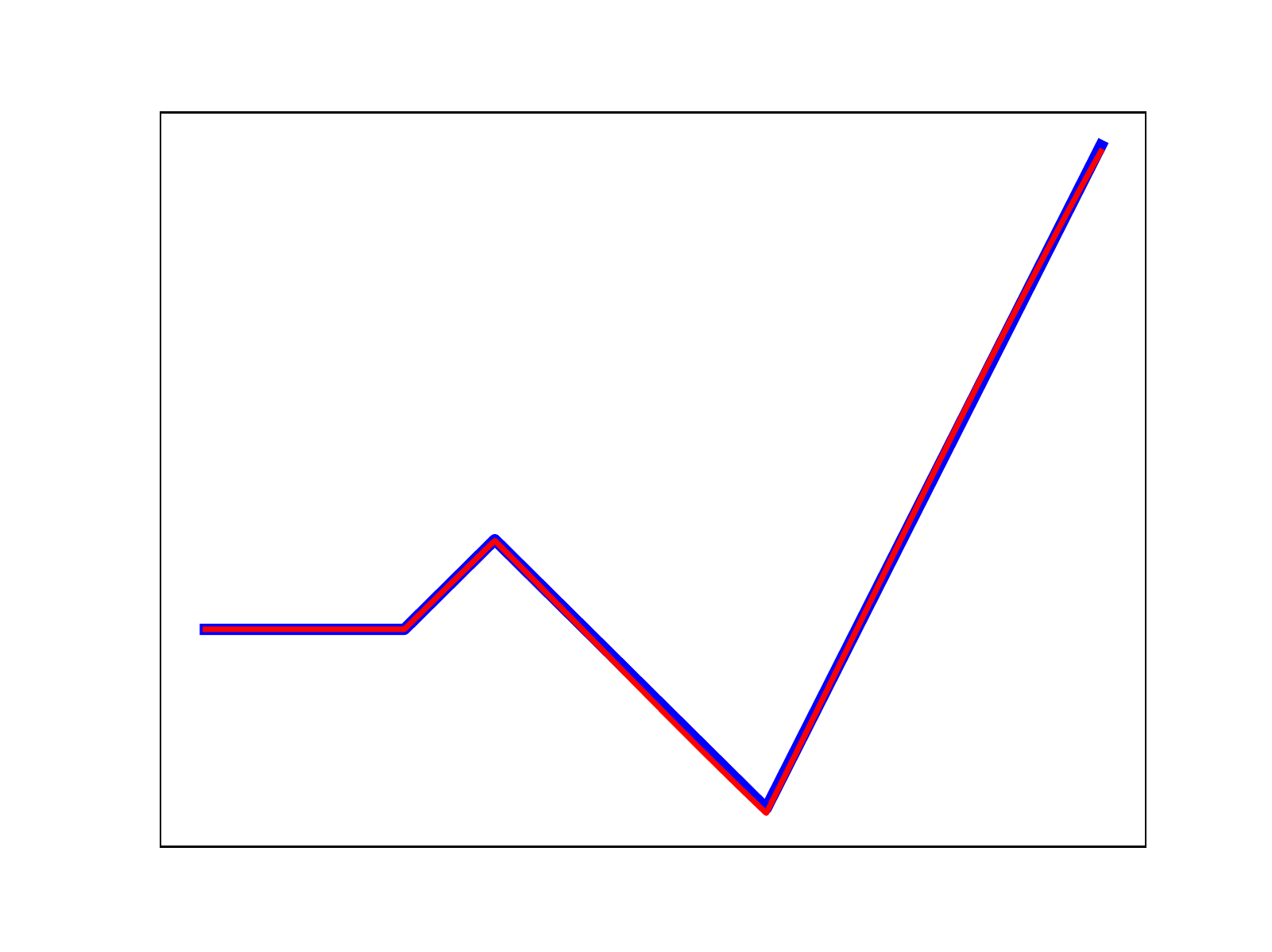}
			\includegraphics[width=0.30\textwidth]{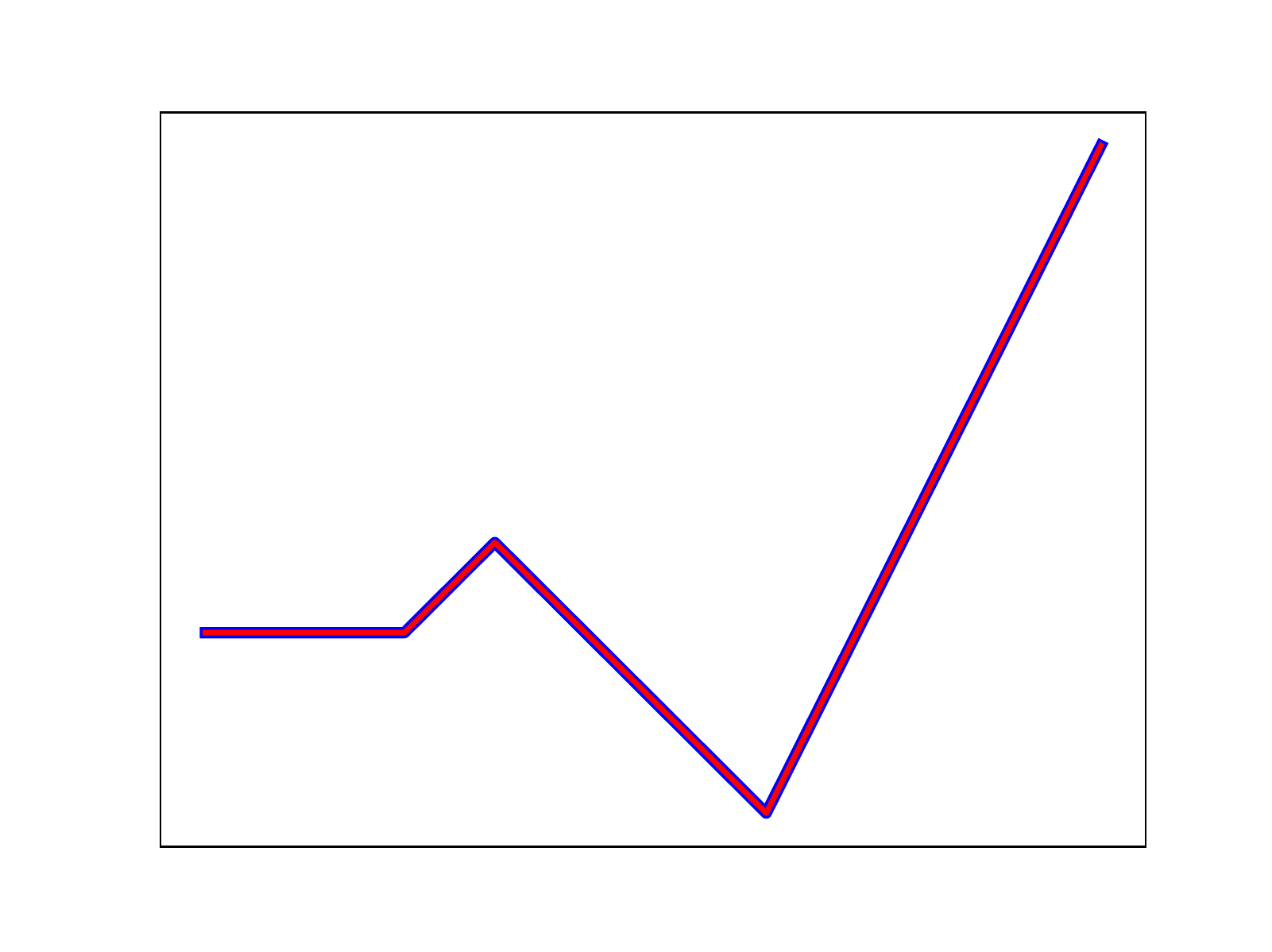}
		\end{subfigure}\vspace{0.3cm}
		\begin{subfigure}[t]{0.9\textwidth}
			\centering
			~~\includegraphics[width=\textwidth]{figures/axis.pdf}
		\end{subfigure}
		\caption{Typical batch SGD}
	\end{subfigure}
	\caption{Examples for rebuild curve from outputs of the experiment with Minibatch SGD and Typical batch SGD. The original curves are in blue and rebuilt curves are in red. The plot shows the outputs of four examples, after 2000, 10000 and 50000 iterations, from left to right. }
	\label{fig:progress}
\end{figure*}

The result of this experiment is presented in Figure \ref{fig:synthetic}. In this benchmark, we can see that typical batch SGD converges faster than conventional Minibatch SGD in terms of both training loss and validation loss. This implies that our proposed typicality sampling scheme is more efficient than SRS and reduces error of gradient estimation while improving generalization capacity. To show this result more explicit, we also plot four curves rebuilt from the outputs after 2000, 10000 and 50000 iterations in Figure \ref{fig:progress}. Clearly, convergence to an accurate solution is faster in typical batch SGD, especially at the early stage of training. The result also demonstrates that Adam, as a complex variant of Minibatch SGD, can also benefit from our special data-driven design of batch selection scheme.

\subsection{Experiments on Natural Dataset}

\begin{figure*}[!t]
	\centering
	\begin{subfigure}[t]{0.32\textwidth}
		\includegraphics[width=\textwidth]{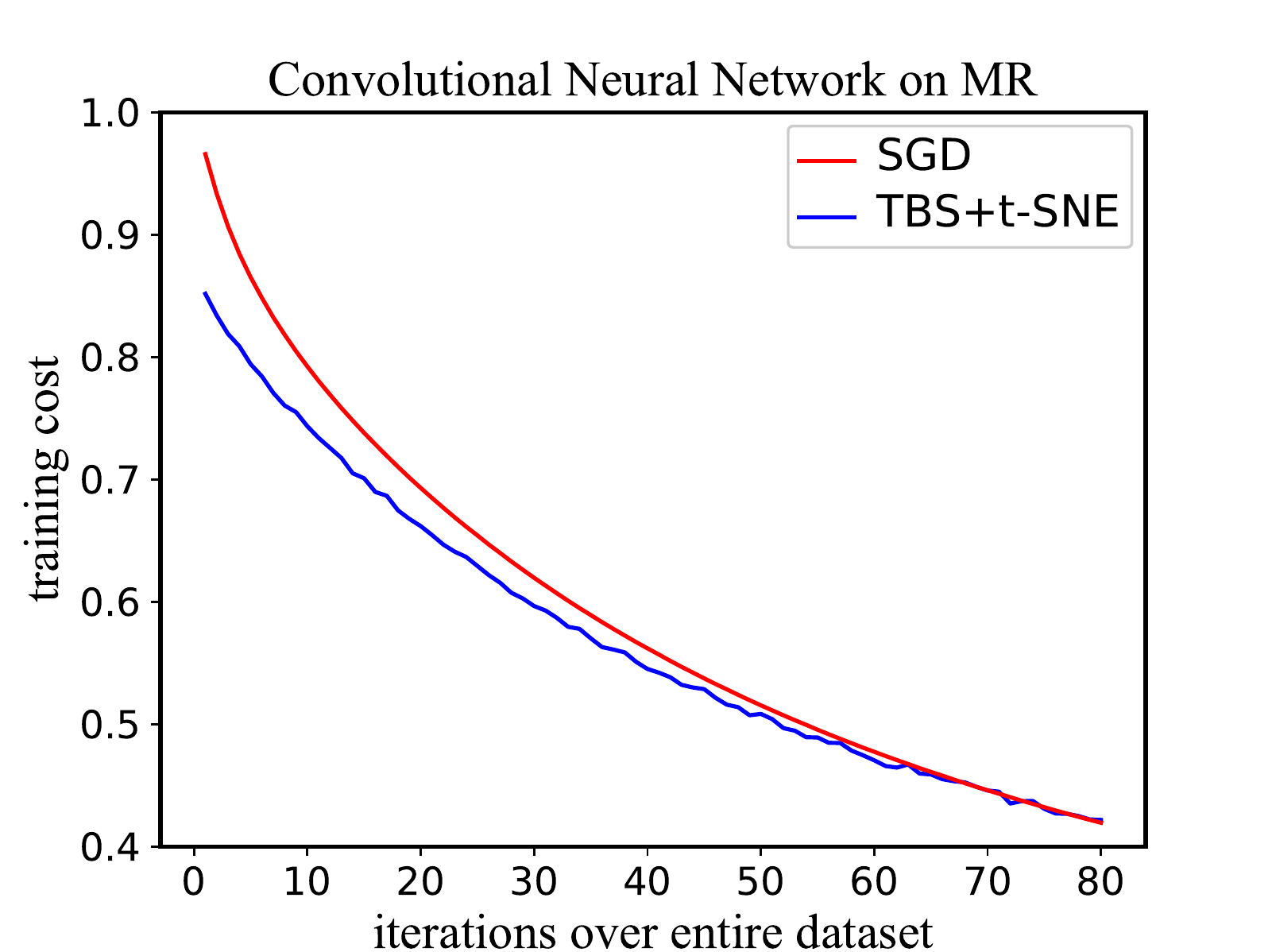}
	\end{subfigure}
	\hfill
	\begin{subfigure}[t]{0.32\textwidth}
		\includegraphics[width=\textwidth]{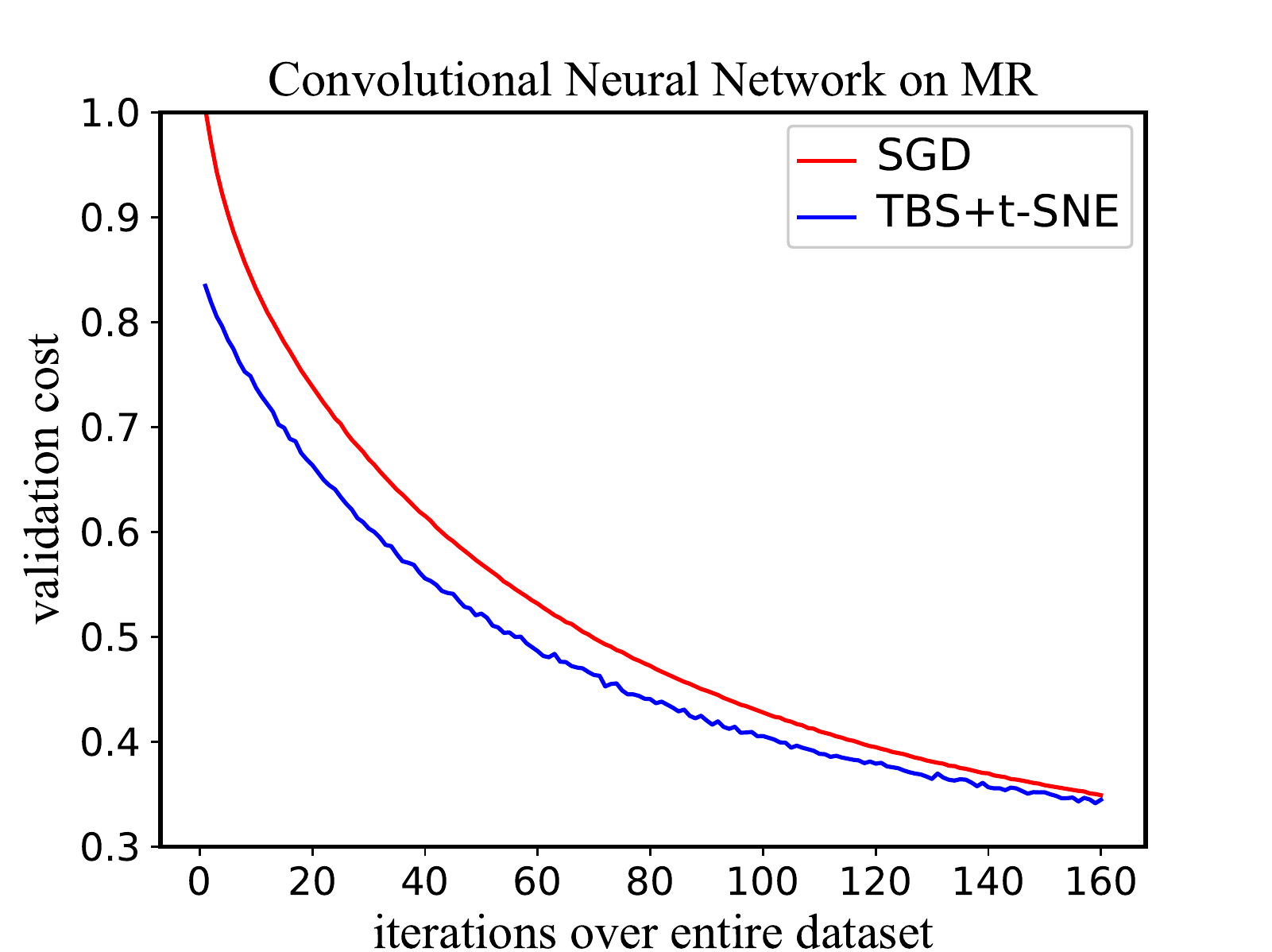}
	\end{subfigure}
	\hfill
	\begin{subfigure}[t]{0.32\textwidth}
		\includegraphics[width=\textwidth]{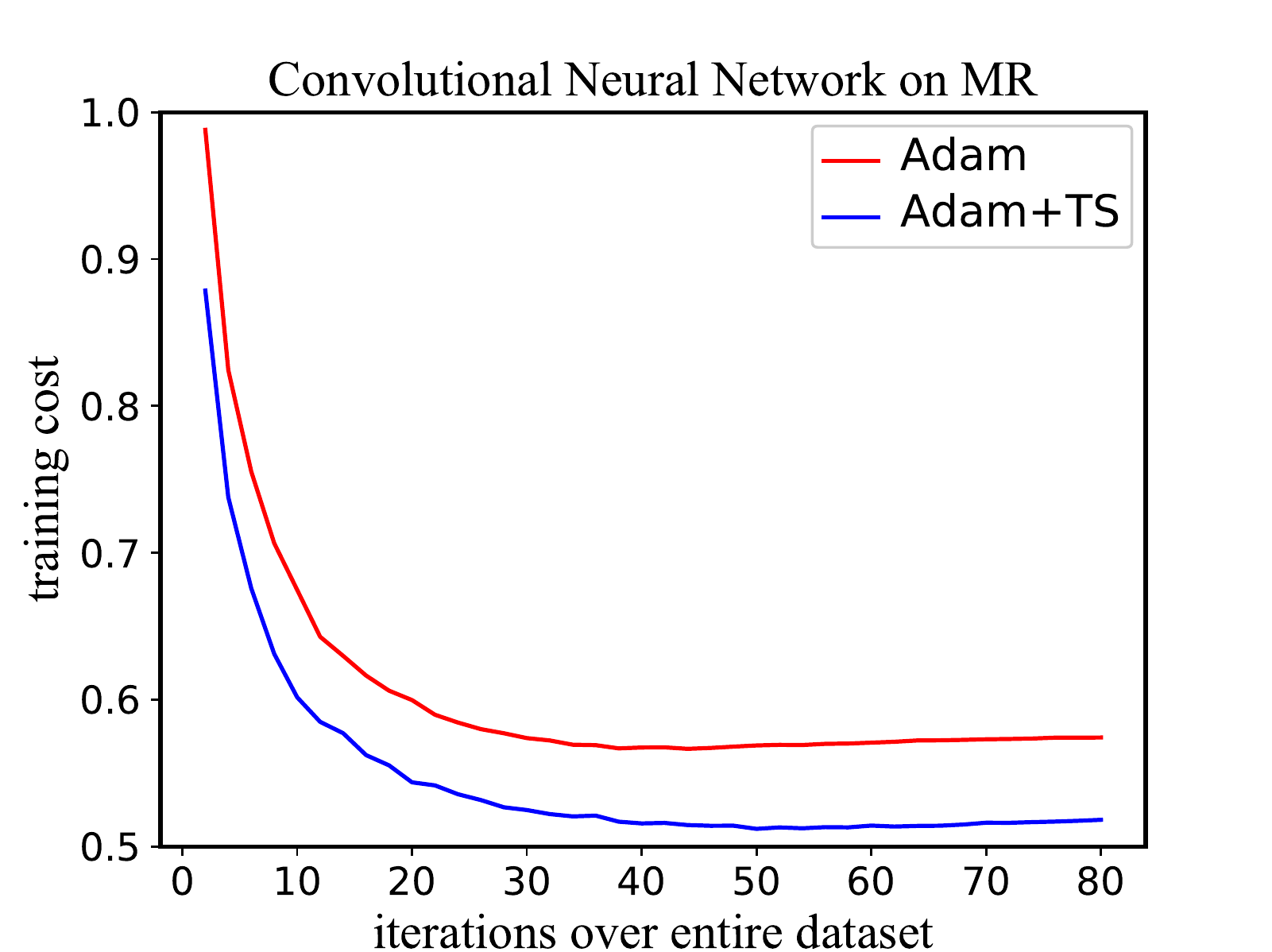}
	\end{subfigure} 
	\hfill
	\begin{subfigure}[t]{0.32\textwidth}
		\includegraphics[width=\textwidth]{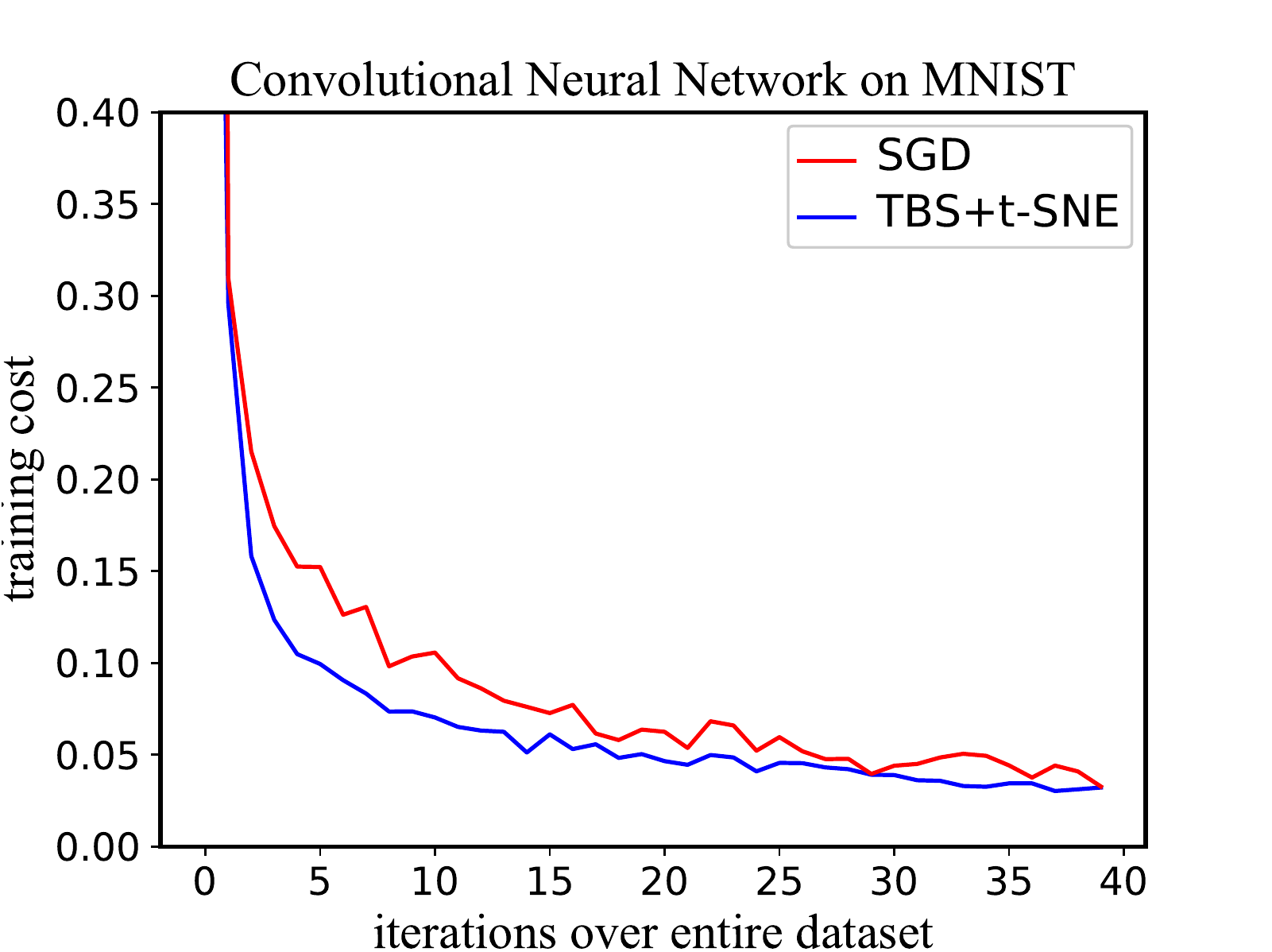}
	\end{subfigure}
	\hfill
	\begin{subfigure}[t]{0.32\textwidth}
		\includegraphics[width=\textwidth]{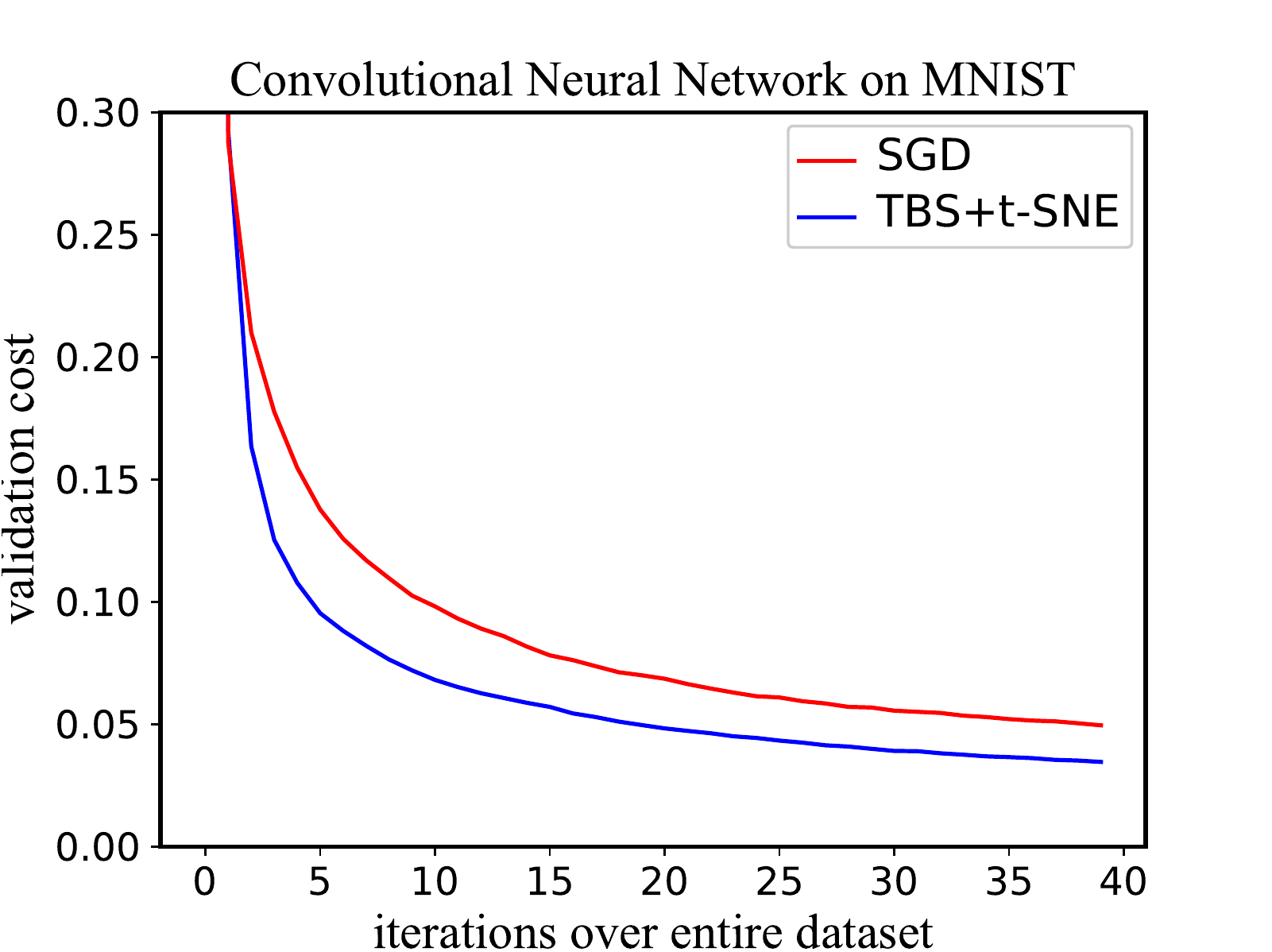}
	\end{subfigure}
	\hfill
	\begin{subfigure}[t]{0.32\textwidth}
		\includegraphics[width=\textwidth]{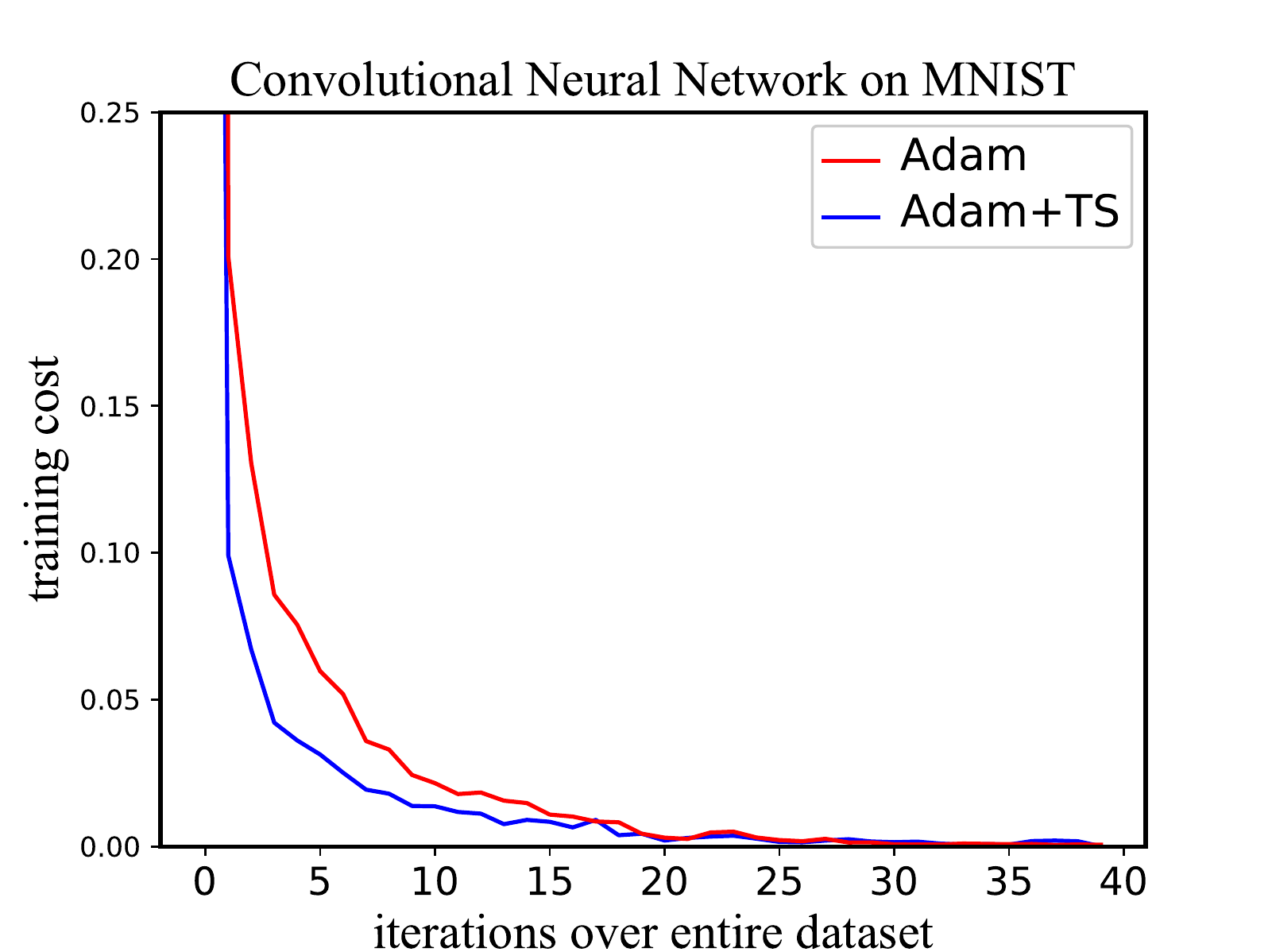}
	\end{subfigure}
	\begin{subfigure}[t]{0.32\textwidth}
		\includegraphics[width=\textwidth]{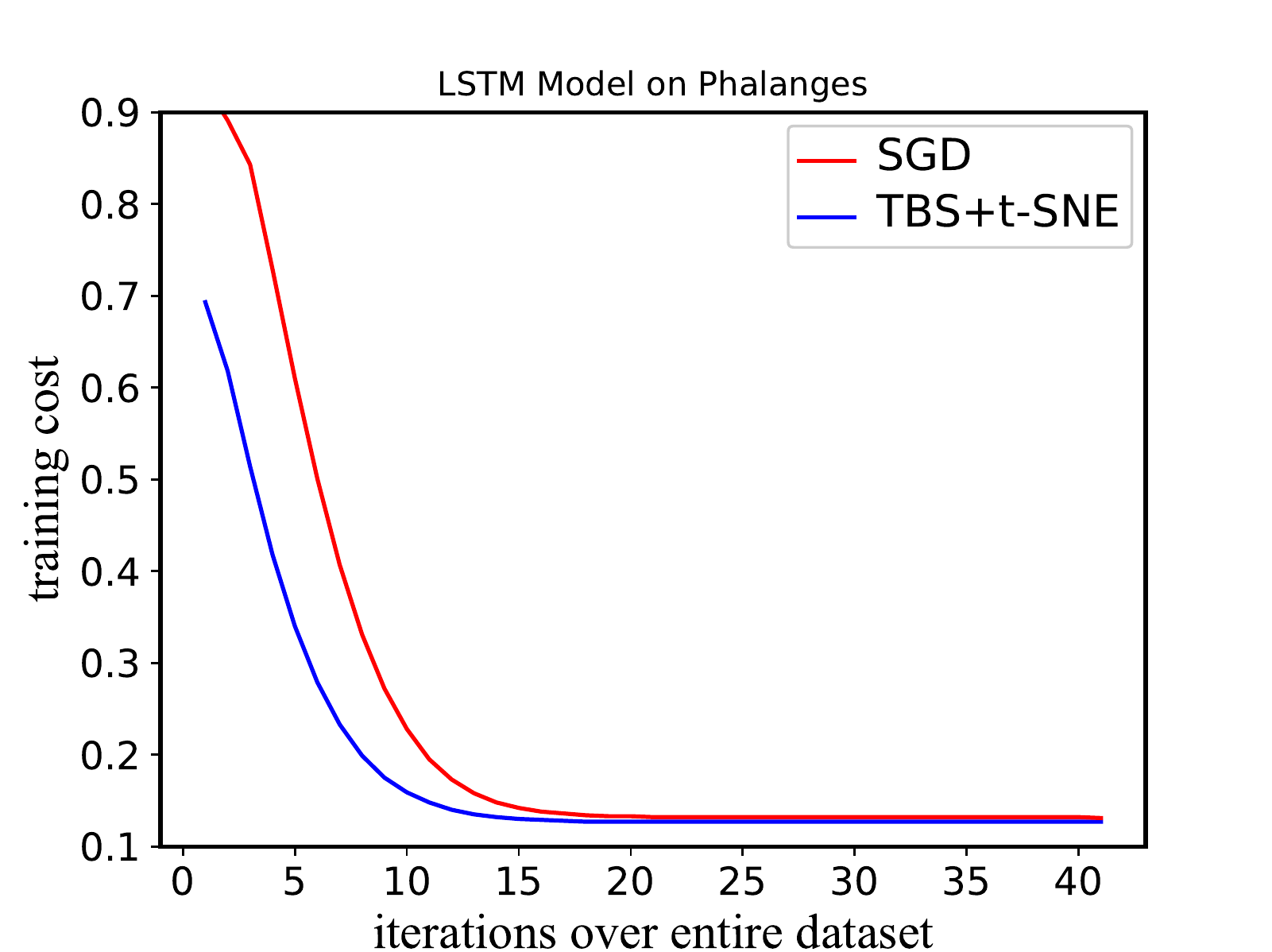}
	\end{subfigure}
	\hfill
	\begin{subfigure}[t]{0.32\textwidth}
		\includegraphics[width=\textwidth]{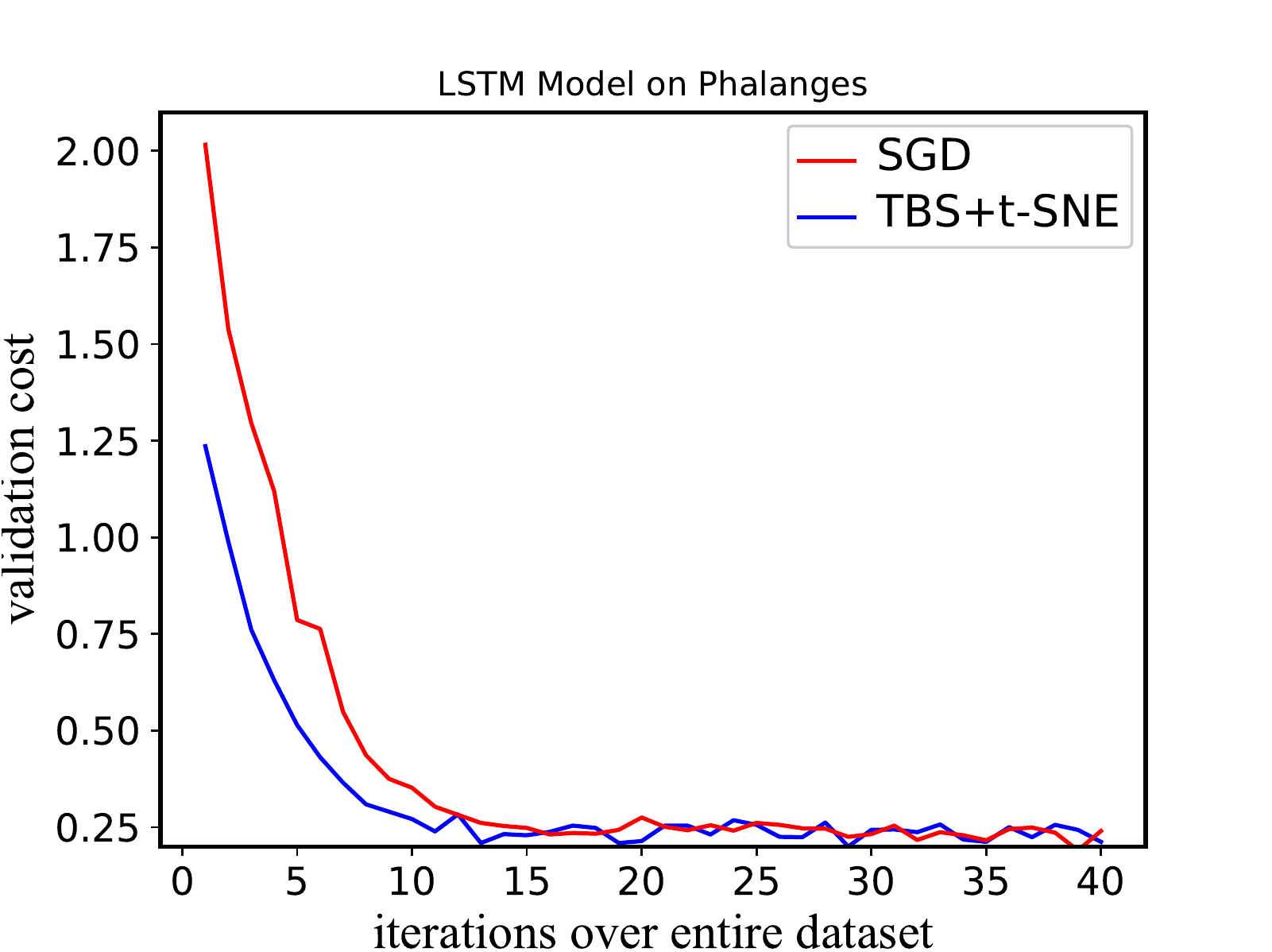}
	\end{subfigure}
	\hfill
	\begin{subfigure}[t]{0.32\textwidth}
		\includegraphics[width=\textwidth]{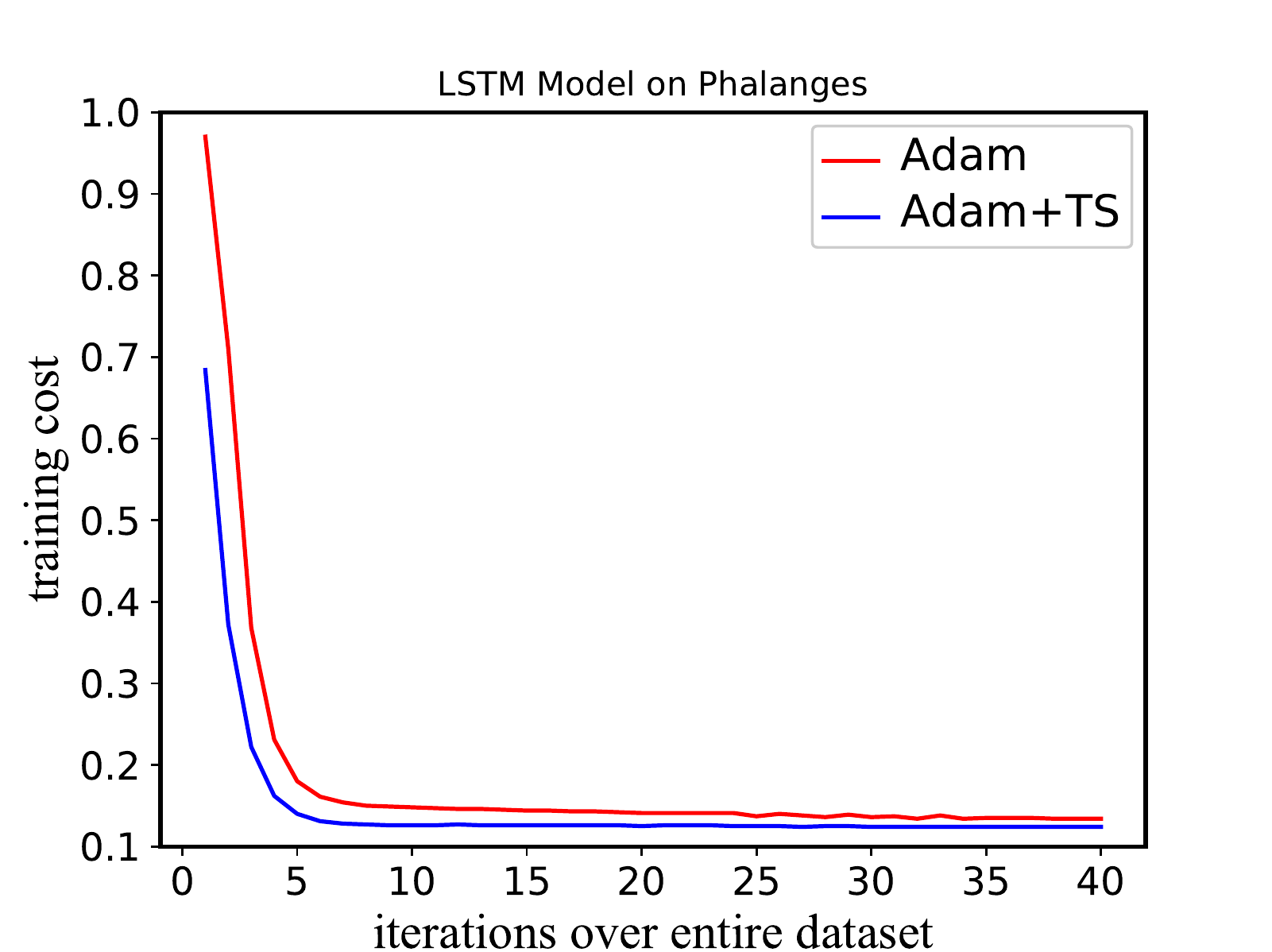}
	\end{subfigure}  
	\caption{Experiments on natural datasets. Top row presents the experimental results for Movie Review dataset, middle row for MNIST dataset and bottom row for Phalanges dataset. The left and middle columns present the curves of training cost and validation cost against Minibatch SGD, the right column present the curves of training cost against Adam.}
	\label{fig:natural}
\end{figure*}

We then evaluate the proposed typicality sampling on natural datasets with more complex models. Three high-dimensional datasets are used, including MNIST dataset \citep{lecun1998gradient} , Movie Review dataset \citep{Pang+Lee:05a} and Phalanges dataset \citep{UCRArchive}. The MNIST dataset consists of 60000 gray-scale handwritten digits images, each has $28*28=784$ pixels. In practice, we select last 5000 images as validation set and keep the rest images as training set. The Movie Review dataset contains 10662 movie reviews with vocabulary of size 20k. Each review is a sentence with average length of 20, which can be classified as positive or negative. The Phalanges dataset is a time series dataset from UCR Archive with a standard validation set. All 2738 time series data are of equal length and the Euclidean distance is given. 

We first train two different CNN models on MNIST dataset for image recognization problem and MR dataset for sentence classification problem. Our CNN architecture tested on MNIST has a 784-dimensional input layer and a convolutional layer of $5*5$ filters of depth $32$ with $2*2$ max polling layers with stride of $2$, followed by two fully connected layer consist of $1024$ and $10$ nodes respectively, with a softmax layer on the top. For MR dataset, our CNN model is composed of $4$ layers. The first layer embeds raw sentences into vectors of length $64$ followed by a convolution layer with filter sizes of $3, 4, 5$. The third layer performs max polling and the last layer is fully connected with $3*32=96$ nodes followed by dropout and softmax activation. We then train a Long Short-Term Memory (LSTM) model on Phalanges dataset. The architecture used is a three-layer stacked LSTM with 120 nodes at each layer, followed by a fully connected layer generating outputs with softmax activation.

We train 40 epochs through training set with mini-batch size of 50 for all three problems. The detailed experiment results are shown in Figure \ref{fig:natural}, from which we can see that across all problems, typical batch SGD outperforms conventional Minibatch SGD on both training set and validation set. We also note that Adam with typicality sampling is superior to plain Adam by achieving faster convergence rate on training set, although our convergence analysis does not apply to this case. All results above support our theoretical intuition that by encouraging high representative samples (in our case, it is the samples in high density regions) to selected with greater probabilities, optimization algorithm using minibatching can obtain faster convergence speed.

\section{Conclusion}
\label{section7}
This paper studies accelerating Minibatch SGD by using typicality sampling scheme. We theoretically prove that the error of gradient estimation can be reduced by the proposed method and thus the resulting typical batch SGD achieves faster convergence. Our experiments on synthetic dataset and natural  dataset validate that typical batch SGD converges consistently faster than conventional Minibatch SGD with better generalization ability, and other variants algorithms can also benefit from the new sampling scheme.

Our typicality sampling also opens several directions of future work. The most important one is making similar idea works to cold start problem in data-driven models. Since the typical samples we select can roughly reflect the pattern of whole training set, we can greatly improve the convergence rate by building more accurate search direction, especially when only a few training samples are available.


%

\appendices
\section{Proof of Lemma \ref{lemma3}}
\begin{IEEEproof}
	Proceeding as the proof of Lemma \ref{lemma2} in \cite{lohr2009sampling}, we begin by defining random variable $ Z_i:=\left\{ \begin{aligned} 1  &~~~~ i \in \batch \\ 0 &~~~ otherwise \end{aligned} \right.$. By this definition, we can rewrite the expectation of gradient estimation with new sampling scheme
	\begin{equation}
	\begin{aligned}
	E\left[{\nabla J_\batch (\theta_k)} \right] &= E\left[ \underset{i\in \mathcal{H}}{\mathop \sum }\,{{Z}_{i}}\frac{\nabla {{J}_{i}}( {{\theta }_{k}})}{m}+\underset{i\in \mathcal{L}}{\mathop \sum }\,{{Z}_{i}}\frac{\nabla {{J}_{i}}({{\theta }_{k}} )}{m} \right] \\ &= \underset{i\in \mathcal{H}}{\mathop \sum }\,E[{{Z}_{i}}]\frac{\nabla {{J}_{i}}( {{\theta }_{k}})}{m}+\underset{i\in \mathcal{L}}{\mathop \sum }\,E[{{Z}_{i}}]\frac{\nabla {{J}_{i}}( {{\theta }_{k}})}{m}.\nonumber
	\end{aligned}
	\end{equation}
	From above observation we can see that proposed batch selection method can be regarded as performing SRS method on subset $\mathcal{H}$ and subset $\mathcal{L}$ respectively, which is exactly the third step in the framework of typical batch SGD. Notice that subset $\mathcal{H}$ is independent of subset $\mathcal{L}$, we obtain a similar result
	\begin{equation}
	\label{proof_lemma3_step1}
	\begin{aligned}
	E\left[ {\nabla J_\batch (\theta_k)} \right]&=\underset{i\in \mathcal{H}}{\mathop \sum }\,\frac{{{n}_{1}}}{{{N}_{1}}}\frac{\nabla {{J}_{i}}( {{\theta }_{k}})}{m}+\underset{i\in \mathcal{L}}{\mathop \sum }\,\frac{{{n}_{2}}}{{{N}_{2}}}\frac{\nabla {{J}_{i}}( {{\theta }_{k}} )}{m}\\   &=\underset{i\in \mathcal{H}}{\mathop \sum }\,\frac{{{n}_{1}}}{{{N}_{1}}}\frac{\nabla {{J}_{i}}( {{\theta }_{k}})}{m}=\beta \nabla J( {{\theta }_{k}}),
	\end{aligned}
	\end{equation}
	where the second equation follows from Assumption \ref{assumption4_trainingset}. Now we derive the expression of sum of variance over all dimension of the gradient estimation with typicality sampling scheme.
	\begin{align}
	\label{proof_lemma3_step2}
	&V( {\nabla J_\batch (\theta_k)})\nonumber\\ 
	&= \frac{1}{{{m}^{2}}}V\left( \underset{i\in \mathcal{H}}{\mathop \sum }\,{{Z}_{i}}\nabla {{J}_{i}}( {{\theta }_{k}} )+\underset{i\in \mathcal{L}}{\mathop \sum }\,{{Z}_{i}}\nabla {{J}_{i}}( {{\theta }_{k}} ) \right) \nonumber \\
	&=\frac{1}{{{m}^{2}}}\Bigg\{ Cov\left( \underset{i\in \mathcal{H}}{\mathop \sum }\,{{Z}_{i}}\nabla {{J}_{i}}( {{\theta }_{k}}),\underset{i\in \mathcal{H}}{\mathop \sum }\,{{Z}_{i}}\nabla {{J}_{i}}( {{\theta }_{k}}) \right)\nonumber\\
	&~~~~~~~~+Cov\left( \underset{i\in \mathcal{L}}{\mathop \sum }\,{{Z}_{i}}\nabla {{J}_{i}}( {{\theta }_{k}}),\underset{i\in \mathcal{L}}{\mathop \sum }\,{{Z}_{i}}\nabla {{J}_{i}}( {{\theta }_{k}}) \right) \Bigg\} \nonumber \\
	&=\frac{1}{{{m}^{2}}}\frac{{{n}_{1}}}{{{N}_{1}}\left( {{N}_{1}}-1 \right)}\left( 1-\frac{{{n}_{1}}}{{{N}_{1}}} \right)\nonumber\\
	&~~~~~~~~~~~~~~~~~~~~*\left[ {{N}_{1}}\underset{i\in \mathcal{H}}{\mathop \sum }\,{\|\nabla {{J}_{j}}{{( {{\theta }_{k}})}\|}^{2}}-{\left\| \underset{i\in \mathcal{H}}{\mathop \sum }\,\nabla {{J}_{i}}( {{\theta }_{k}} ) \right\|^{2}} \right] \nonumber \\
	&~~~~~~~+\frac{1}{{{m}^{2}}}\frac{{{n}_{2}}}{{{N}_{2}}\left( {{N}_{2}}-1 \right)}\left( 1-\frac{{{n}_{2}}}{{{N}_{2}}} \right)\nonumber\\
	&~~~~~~~~~~~~~~~~~~~~*\left[ {{N}_{2}}\underset{i\in \mathcal{L}}{\mathop \sum }\,{\|\nabla {{J}_{j}}{{\left( {{\theta }_{k}} \right)}\|}^{2}}-{{\left\| \underset{i\in \mathcal{L}}{\mathop \sum }\,\nabla {{J}_{i}}( {{\theta }_{k}}) \right\|}^{2}} \right] \nonumber \\
	&=\left( 1-\frac{{{n}_{1}}}{{{N}_{1}}} \right)\frac{{{n}_{1}}}{{{m}^{2}}}\frac{1}{{{N}_{1}}-1}\underset{i\in \mathcal{H}}{\mathop \sum }\,{{\left\| \nabla {{J}_{i}}( {{\theta }_{k}} )-\nabla J( {{\theta }_{k}} ) \right\|}^{2}}\nonumber\\
	&~~~~~~~+\left( 1-\frac{{{n}_{2}}}{{{N}_{2}}} \right)\frac{{{n}_{2}}}{{{m}^{2}}}\frac{1}{{{N}_{2}}-1}\underset{i\in \mathcal{L}}{\mathop \sum }\,{\|\nabla {{J}_{i}}{{( {{\theta }_{k}} )}\|}^{2}}\nonumber\\
	&=\left( 1-\frac{{{n}_{1}}}{{{N}_{1}}} \right)\frac{{{n}_{1}}}{{{m}^{2}}}\mathcal{S}_{H,k}^{2}+\left( 1-\frac{{{n}_{2}}}{{{N}_{2}}} \right)\frac{{{n}_{2}}}{{{m}^{2}}}\mathcal{S}_{\ell,k}^{2},
	\end{align}
	where the first and second equations use the definition of variance, the third equation performances the intermediate result of Lemma \ref{lemma2}  on both subset $\mathcal{H}$ and subset $\mathcal{L}$, and the forth equation uses the conditions in Assumption \ref{assumption4_trainingset}. Combining \eqref{proof_lemma3_step1} and \eqref{proof_lemma3_step2} with the fact $E{{\| {{e}_{k}} \|}^{2}} = E{{\|{\nabla J_\batch (\theta_k)}-\nabla J( {{\theta }_{k}} ) \|}^{2}} = V({\nabla J_\batch (\theta_k)})+{{\left\|E[ {\nabla J_\batch (\theta_k)}] - \nabla J({{\theta }_{k}}) \right\|}^{2}}$ yields 
    \begin{eqnarray}
    E{{\|{{e}_{k}} \|}^{2}}&=&{{\Big\|\big(\beta-1\big)\cdot\nabla J( {{\theta   }_{k}})\Big\|}^{2}}+(1-\frac{{n}_{1}}{{{N}_{1}}})\frac{n_1}{m^2}\mathcal{S}_{H,k}^{2} \nonumber\\
    & &+(1-\frac{n_2}{{{N}_{2}}} )\frac{n_2}{m^2}\mathcal{S}_{\ell,k}^{2}\nonumber
    \end{eqnarray}
	The result follows immediately.	
\end{IEEEproof}

\section{Proof of Theorem \ref{convergence_rate}}
\begin{IEEEproof}
	We begin by deriving an upper bound on expectation of gradient error. From Lemma \ref{lemma2}, we apply triangle inequality on the result
	\begin{align}
	&E{{\|{{e}_{k}} \|}^{2}} \le {{\Big\| \left( \beta -1 \right)\cdot \nabla J( {{\theta }_{k}}) \Big\|}^{2}}\nonumber\\
	&~~~~~~~~~~~~~+\left( 1-\frac{n_1}{{{N}_{1}}} \right)\frac{n_1}{m^2}\frac{{{N}_{1}}}{{{N}_{1}}-1}*\left( 2{\|\nabla {{J}_{i}}{{( {{\theta }_{k}} )}\|}^{2}}+ 2{\|\nabla J{{( {{\theta }_{k}})}\|}^{2}} \right)\nonumber\\
	&~~~~~~~~~~~~~+\left( 1-\frac{n_2}{{{N}_{2}}} \right)\frac{n_2}{m^2}\frac{{{N}_{2}}}{{{N}_{2}}-1}{\|\nabla {{J}_{i}}{{( {{\theta }_{k}})}}\|}^{2}.\nonumber
	\end{align}
	Combine Assumption \ref{assumption3_stochistic_ek} to substitute $\|\nabla {{J}_{i}}{{( {{\theta }_{k}} )}\|}^{2}$ in above inequality 
	\begin{equation}
	\label{theorem1-step2}
	\begin{aligned}
	&E{{\|{{e}_{k}} \|}^{2}} \le {{\Big\| \left( \beta -1 \right)\cdot\nabla J( {{\theta }_{k}} ) \Big\|}^{2}}\\
	&~~~~~~~~~~~~~~+\left( 1-\frac{n_1}{{{N}_{1}}} \right)\frac{n_1}{m^2}\frac{{{N}_{1}}}{{{N}_{1}}-1}
	*\left( 2{{\beta }_{2}}+4 \right){\|\nabla J{{( {{\theta }_{k}})}\|}^{2}}\\
	&~~~~~~~~~~~~~~+\left( 1-\frac{n_2}{{{N}_{2}}} \right)\frac{n_2}{m^2}\frac{{{N}_{2}}}{{{N}_{2}}-1}\left( {{\beta }_{2}}+1 \right){\|\nabla J{{( {{\theta }_{k}} )}\|}^{2}}.
	\end{aligned}
	\end{equation}
	We can also get following result from Assumption \ref{assumption1_lipschitz}
	\begin{equation}
	{{\| \nabla J( {{\theta }_{k}}) \|}^{2}}\le 2L \left( J( {{\theta }_{k}})-J( {{\theta }_{*}} ) \right).\nonumber
	\end{equation}
	Thus the bound on $E{{\|{{e}_{k}} \|}^{2}}$ in \eqref{theorem1-step2} can be expressed in terms of the distance to optimality $J( {{\theta }_{*}} )$
	\begin{align}
	\label{theorem1-step4}
	\begin{split}
	E{{\|{{e}_{k}} \|}^{2}} &\le \Bigg\{ {{\left( \beta -1 \right)}^{2}}+\left( 1-\frac{n_1}{{{N}_{1}}} \right)\frac{n_1}{m^2}\frac{{{N}_{1}}}{{{N}_{1}}-1}\left( 2{{\beta }_{2}}+4 \right)\\
	&~~~~+\left( 1-\frac{n_2}{{{N}_{2}}} \right)\frac{n_2}{m^2}\frac{{{N}_{2}}}{{{N}_{2}}-1}\left( {{\beta }_{2}}+1 \right) \Bigg\}*2L\left( J( {{\theta }_{k}})-J( {{\theta }_{*}} ) \right)
	\end{split}
	\end{align}
	Using \eqref{theorem1-step4} to replace $E{{\|{{e}_{k}} \|}^{2}}$ in Lemma \ref{lemma1} can obtain
	\begin{equation}
	\begin{split}
	E&[J({{\theta }_{k+1}} )-J( {{\theta }_{*}} ) ]\le \Big[ 1-\frac{\mu }{L}+( 1-\frac{n_1}{{{N}_{1}}} )\frac{2n_1(\beta_{2}+2)}{m^2}
	\\&+( 1-\frac{n_2}{{{N}_{2}}} )\frac{n_2(\beta_{2}+1)}{2m^2}+{{( \beta -1 )}^{2}} \Big]*E[ ( J( {{\theta }_{k}} )-J( {{\theta }_{*}} ) ) ]\nonumber
	\end{split}
	\end{equation}		
\end{IEEEproof}

\section{Proof of Theorem \ref{compare_with_sgd}}
\begin{IEEEproof}
	From Lemma \ref{lemma2} and Lemma \ref{lemma3} we have
	\begin{equation}
	\begin{aligned}
	E{{\| {{e}_{k}} \|}^{2}}_{ori}&-E{{\| {{e}_{k}} \|}^{2}}=\left( 1-\frac{m}{N} \right)\frac{\mathcal{S}_{k}^{2}}{m}-{\|{{\left( \beta -1 \right)}}\cdot\nabla J{{( {{\theta }_{k}})}}\|}^{2}\nonumber\\
	&-\left( 1-\frac{n_1}{{{N}_{1}}} \right)\frac{n_1}{m^2}\mathcal{S}_{H,k}^{2}-\left( 1-\frac{n_2}{{{N}_{2}}} \right)\frac{n_2}{m^2}\mathcal{S}_{\ell,k}^{2}\nonumber\\
	\end{aligned}
	\end{equation}
	Expand the right side of above equation and use Assumption \ref{assumption4_trainingset} to eliminate $\underset{i\in \mathcal{L}}{\mathop \sum }\,{\nabla {{J}_{i}}{{( {{\theta }_{k}})}}}$
	\begin{equation}
	\begin{aligned}
	&E{{\| {{e}_{k}} \|}^{2}}_{ori}-E{{\| {{e}_{k}} \|}^{2}}\nonumber\\
	&=\frac{1}{{{m}^{2}}}\left[ \left( 1-\frac{m}{N} \right)\frac{m}{N}-\left( 1-\frac{n_2}{{{N}_{2}}} \right)\frac{n_2}{{{N}_{2}}} \right]\underset{i\in \mathcal{L}}{\mathop \sum }\,{\|\nabla {{J}_{i}}{( {{\theta }_{k}})}\|}^{2}\nonumber\\
	&~~~~~~~+\left[ \left( 1-\frac{m}{N} \right)\left( \frac{1}{m} \right)\left( \frac{{{N}_{2}}}{N} \right)-{{\left( \beta -1 \right)}^{2}} \right]{\|\nabla J{\left( {{\theta }_{k}} \right)}\|}^{2}\nonumber\\
	&+\frac{1}{{{m}^{2}}}\left[ \left( 1-\frac{m}{N} \right)\frac{m}{N}-\left( 1-\frac{n_1}{{{N}_{1}}} \right)\frac{n_1}{{{N}_{1}}} \right]\underset{i\in \mathcal{H}}{\mathop \sum }\,{{\left\| \nabla {{J}_{i}}( {{\theta }_{k}})-\nabla J( {{\theta }_{k}} \right)\|}^{2}}\nonumber\\
	&=\frac{1}{{{m}^{2}}}\left[ \left( 1-\frac{m}{N} \right)\frac{m}{N}-\left( 1-\frac{n_1}{{{N}_{1}}} \right)\frac{n_1}{{{N}_{1}}} \right]\underset{i\in \mathcal{H}}{\mathop \sum }\,\nabla {\|{{J}_{i}}{{( {{\theta }_{k}} )}\|}^{2}}\nonumber\\	
	&~~~~~~~+\left[ \left( 1-\frac{m}{N} \right)\left( \frac{1}{m} \right)\left( \frac{{{N}_{2}}}{N} \right)-{{\left( \beta -1 \right)}^{2}} \right]{\|\nabla J{{( {{\theta }_{k}})}\|}^{2}}\nonumber\\
	&~~~~~~~+\frac{1}{{{m}^{2}}}\left[ \left( 1-\frac{m}{N} \right)\frac{m}{N}-\left( 1-\frac{n_2}{{{N}_{2}}} \right)\frac{n_2}{{{N}_{2}}} \right]\underset{i\in \mathcal{L}}{\mathop \sum }\,{\|\nabla {{J}_{i}}{{( {{\theta }_{k}})}\|}^{2}}\nonumber\\
	&~~~~~~~+\frac{1}{{{m}^{2}}}\left[ \left( 1-\frac{n_1}{{{N}_{1}}} \right)\frac{n_1}{{{N}_{1}}}\underset{i\in \mathcal{H}}{\mathop \sum }\,{\|\nabla J{{( {{\theta }_{k}})}\|}^{2}} \right]\nonumber\\
	&~~~~~~~-\frac{1}{{{m}^{2}}}\left[ \left( 1-\frac{m}{N} \right)\frac{m}{N}\underset{i\in \mathcal{H}}{\mathop \sum }\,{\|\nabla J{{\left( {{\theta }_{k}} \right)}\|}^{2}} \right]\nonumber
	\end{aligned}
	\end{equation}
	Using \eqref{size_of_n_1} and Assumption \ref{assumption3_stochistic_ek} to substitute $\|\nabla {{J}_{i}}{{( {{\theta }_{k}} )}\|}^{2}$, we can get
	\begin{equation}
	\begin{aligned}
	&E{{\| {{e}_{k}} \|}^{2}}_{ori}-E{{\| {{e}_{k}} \|}^{2}}\nonumber\\
	&\ge \frac{1}{{{m}^{2}}}\left[ \left( 1-\frac{m}{N} \right)\frac{m}{N}-\left( 1-\frac{n_2}{{{N}_{2}}} \right)\frac{n_2}{{{N}_{2}}} \right]\underset{i\in \mathcal{L}}{\mathop \sum }\,{{\beta }_{2}}{\|\nabla J{{( {{\theta }_{k}})}\|}^{2}}\nonumber\\
	&~~~~~~~+\left[ \left( 1-\frac{m}{N} \right)\left( \frac{1}{m} \right)\left( \frac{{{N}_{2}}}{N} \right)-{{\left( \beta -1 \right)}^{2}} \right]{\|\nabla J{{( {{\theta }_{k}} )}\|}^{2}}\nonumber\\
	&~~~~~~~+\frac{1}{{{m}^{2}}}\left[ \left( 1-\frac{n_1}{{{N}_{1}}} \right)\frac{n_1}{{{N}_{1}}}\underset{i\in \mathcal{H}}{\mathop \sum }\,{\|\nabla J{{( {{\theta }_{k}} )}\|}^{2}} \right]\nonumber\\
	&~~~~~~~-\frac{1}{{{m}^{2}}}\left[ \left( 1-\frac{m}{N} \right)\frac{m}{N}\underset{i\in \mathcal{H}}{\mathop \sum }\,{\|\nabla J{{( {{\theta }_{k}})}\|}^{2}} \right]\nonumber\\
	&+\frac{1}{{{m}^{2}}}\left[ \left( 1-\frac{m}{N} \right)\frac{m}{N}-\left( 1-\frac{n_1}{{{N}_{1}}} \right)\frac{n_1}{{{N}_{1}}} \right]\underset{i\in \mathcal{H}}{\mathop \sum }\,\left({{\beta }_{1}}+{{\beta }_{2}}{\|\nabla J{{( {{\theta }_{k}})}\|}^{2}}\right)\nonumber\\
	\end{aligned}
	\end{equation}
	In addition, it follows from Assumption \ref{assumption3_stochistic_ek} that ${\|\nabla J{{( {{\theta }_{k}})}\|}^{2}}\ge\beta_{1}$, thus
	\begin{equation}
	\begin{aligned}
	&E{{\| {{e}_{k}} \|}^{2}}_{ori}-E{{\| {{e}_{k}} \|}^{2}}\nonumber\\
	&\ge \frac{1}{{{m}^{2}}}\left[ \left( 1-\frac{m}{N} \right)\frac{m}{N}-\left( 1-\frac{n_2}{{{N}_{2}}} \right)\frac{n_2}{{{N}_{2}}} \right]\underset{i\in \mathcal{L}}{\mathop \sum }\,{{\beta }_{2}}{\|\nabla J{{( {{\theta }_{k}})}\|}^{2}}\nonumber\\
	&~~~~~~~+\left[ \left( 1-\frac{m}{N} \right)\left( \frac{1}{m} \right)\left( \frac{{{N}_{2}}}{N} \right)-{{\left( \beta -1 \right)}^{2}} \right]{\|\nabla J{{( {{\theta }_{k}} )}\|}^{2}}\nonumber\\
	&~~~~~~~+\frac{1}{{{m}^{2}}}\left[ \left( 1-\frac{m}{N} \right)\frac{m}{N}-\left( 1-\frac{n_1}{{{N}_{1}}} \right)\frac{n_1}{{{N}_{1}}} \right]\underset{i\in \mathcal{H}}{\mathop \sum }\,{{\beta }_{2}}{\|\nabla J{{( {{\theta }_{k}})}\|}^{2}}\nonumber\\
	& = \frac{{\beta }_{2}{\|\nabla J{{( {{\theta }_{k}})}\|}^{2}}}{{{m}^{2}}}\left[\left(1-\frac{m}{N}\right) m - \Big(1-\frac{n_1}{N_1}\Big) n_1-\Big(1-\frac{n_2}{N_2}\Big) n_2\right]\nonumber\\
	&~~~~~~~+\left[ \left( 1-\frac{m}{N} \right)\Big( \frac{1}{m} \Big)\Big( \frac{{{N}_{2}}}{N} \Big)-{{\left( \beta -1 \right)}^{2}} \right]{\|\nabla J{{( {{\theta }_{k}} )}\|}^{2}}\nonumber\\
	\end{aligned}
	\end{equation}
	Rearranging above inequality can obtain, after simplifying
	\begin{equation}
	\begin{aligned}
	&~E{{\| {{e}_{k}} \|}^{2}}_{ori}-E{{\| {{e}_{k}} \|}^{2}}\nonumber\\
	&\ge \frac{{\beta }_{2}{\|\nabla J{{( {{\theta }_{k}})}\|}^{2}}}{{{m}^{2}}}\left[\frac{1}{N}\left(\sqrt{\frac{N_2}{N_1}n_1}-\sqrt{\frac{N_1}{N_2}n_2}\right)^{2}\right]\nonumber\\
	&~~~~~~~+\left[ \left( 1-\frac{m}{N} \right)\Big( \frac{1}{m} \Big)\Big( \frac{{{N}_{2}}}{N} \Big)-{{\left( \beta -1 \right)}^{2}} \right]{\|\nabla J{{( {{\theta }_{k}} )}\|}^{2}}\nonumber\\\\
	& \ge 0,
	\end{aligned}
	\end{equation}
	which proves the theorem.
\end{IEEEproof}

\ifCLASSOPTIONcaptionsoff
  \newpage
\fi

\bibliographystyle{IEEEtran}
\bibliography{arxiv-typ-sampling}

\end{document}